\documentclass[acmtog]{acmart}

\usepackage{multirow}
\usepackage{bbding}
\usepackage{ulem}

\AtBeginDocument{%
  }

\setcopyright{acmlicensed}
\copyrightyear{2025}
\acmYear{2025}
\acmDOI{XXXXXXX.XXXXXXX}

\newcommand{\ghl}[1]{\textcolor{purple}{#1}}

\newcommand{\drawred}[1]{\textcolor{red}{#1}}
\newcommand{\drawgreen}[1]{\textcolor{green}{#1}}

\acmConference[Preprint]{Make sure to enter the correct conference title from your rights confirmation emai}{2025}{Any2AnyTryon report}

\begin{document}

\title{Any2AnyTryon: Leveraging Adaptive Position Embeddings for Versatile Virtual Clothing Tasks}



\author{Hailong Guo}
\authornote{This work was completed during the authors' internship at Tiamat AI.}
\email{guohailong@bupt.edu.cn}
\affiliation{%
  \institution{Beijing University of Posts and Telecommunications}
  \city{Beijing}
  \country{China}}

\author{Bohan Zeng}
\authornotemark[1]
\affiliation{%
  \institution{Peking University}
  \city{Beijing}
  \country{China}}

\author{Yiren Song}
\authornotemark[1]
\affiliation{%
  \institution{National University of Singapore}
  \country{Singapore}}

\author{Wentao Zhang}
\affiliation{%
  \institution{Peking University}
  \city{Beijing}
  \country{China}}

\author{Chuang Zhang}
\authornote{Corresponding Author}
\affiliation{%
  \institution{Beijing University of Posts and Telecommunications}
  \city{Beijing}
  \country{China}}
\email{zhangchuang@bupt.edu.cn}

\author{Jiaming Liu}
\authornote{Project Leader}
\affiliation{%
 \institution{TiamatAI}
 \city{Beijing}
  \country{China}}
\email{jmliu1217@gmail.com}

\renewcommand{\shortauthors}{Guo et al.}

\begin{abstract}
Image-based virtual try-on (VTON) aims to generate a virtual try-on result by transferring an input garment onto a target person’s image. However, the scarcity of paired garment-model data makes it challenging for existing methods to achieve high generalization and quality in VTON. Also, it limits the ability to generate mask-free try-ons. To tackle the data scarcity problem, approaches such as Stable Garment and MMTryon use a synthetic data strategy, effectively increasing the amount of paired data on the model side. However, existing methods are typically limited to performing specific try-on tasks and lack user-friendliness.

To enhance the generalization and controllability of VTON generation, we propose Any2AnyTryon, which can generate try-on results based on different textual instructions and model garment images to meet various needs, eliminating the reliance on masks, poses, or other conditions. Specifically, we first construct the virtual try-on dataset LAION-Garment, the largest known open-source garment try-on dataset. Then, we introduce adaptive position embedding, which enables the model to generate satisfactory outfitted model images or garment images based on input images of different sizes and categories, significantly enhancing the generalization and controllability of VTON generation. In our experiments, we demonstrate the effectiveness of our Any2AnyTryon and compare it with existing methods. The results show that Any2AnyTryon enables flexible, controllable, and high-quality image-based virtual try-on generation.
\ghl{\url{https://logn-2024.github.io/Any2anyTryonProjectPage/}}
\end{abstract}

\begin{teaserfigure}
  \centering
  \includegraphics[width=0.84\textwidth]{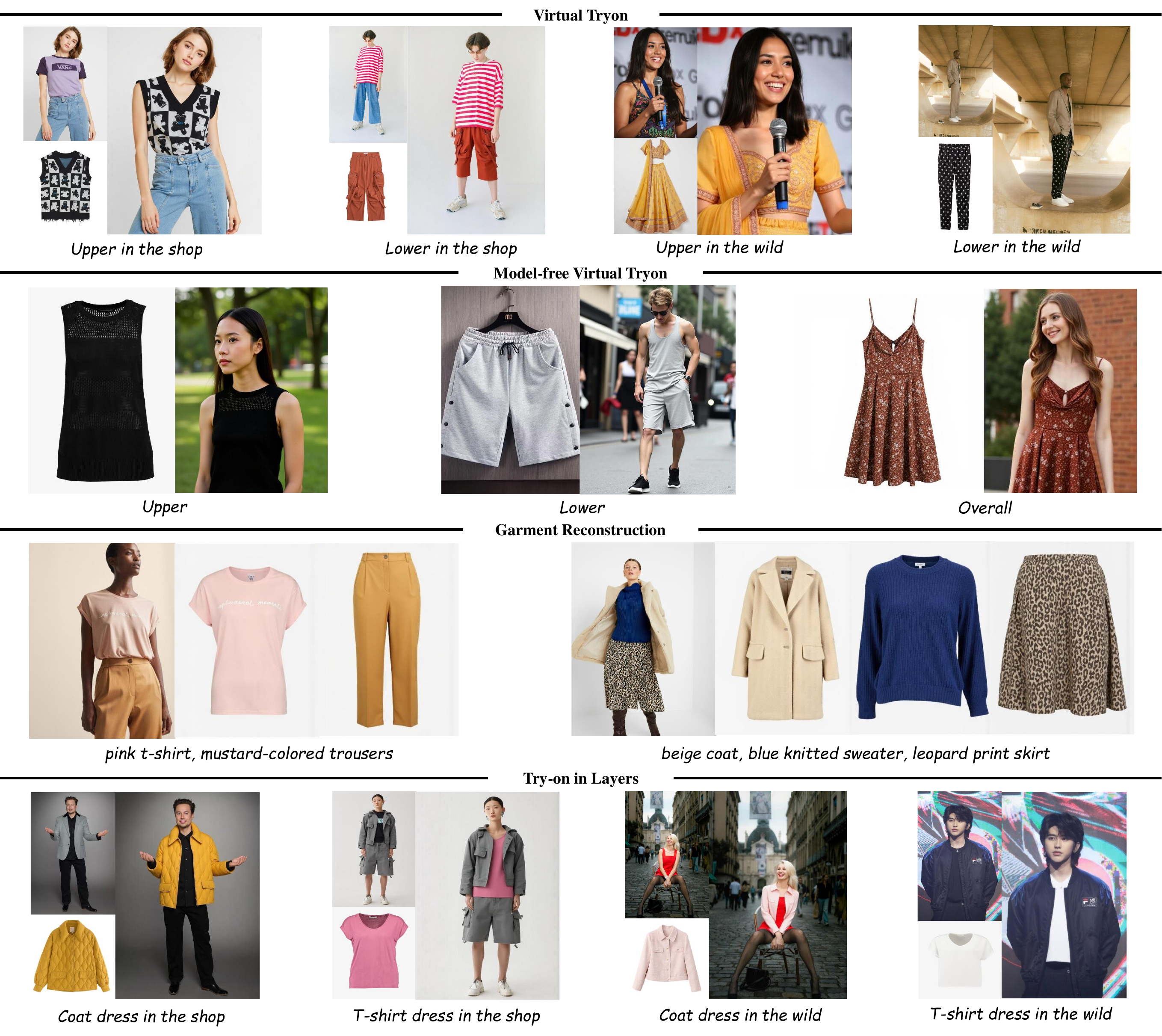}
  \caption{Outfitted model and garment images generated by Any2AnyTryon.}
  \label{fig:teaser}
\end{teaserfigure}


\maketitle

\begin{table*}[!h]
  \small
  \centering
  \setlength{\abovecaptionskip}{0cm}
  \caption{Comparison of VTON functionalities achieved by our Any2AnyTryon and previous methods.}
  \label{Tab:method_func_compare}
  \vspace{0.2cm}
\begin{tabular}{lcccccccc}
  \toprule
  \textbf{Method} & Virtual tryon  & Model-free virtual tryon  & Garment reconstruction   & Try-on in layers  \\
  \midrule
  DiOr~\cite{cui2021dressing}  & \drawgreen{\Checkmark}   & \drawred{\XSolidBrush}   & \drawred{\XSolidBrush}   &  \drawgreen{\Checkmark}  \\
  TryOffDiff~\cite{velioglu2024tryoffdiff}  & \drawred{\XSolidBrush}   & \drawred{\XSolidBrush}   & \drawgreen{\Checkmark}   & \drawred{\XSolidBrush}   \\
  StableGarment~\cite{wang2024stablegarment}  &  \drawgreen{\Checkmark}  &  \drawgreen{\Checkmark}  & \drawred{\XSolidBrush}   & \drawred{\XSolidBrush}   \\
  Magic Clothing~\cite{chen2024magic}  & \drawred{\XSolidBrush}   &    \drawgreen{\Checkmark} & \drawred{\XSolidBrush}   & \drawred{\XSolidBrush}   \\
  GP-VTON~\cite{xie2023gp}  & \drawgreen{\Checkmark}   & \drawred{\XSolidBrush}   & \drawred{\XSolidBrush}   & \drawred{\XSolidBrush}   \\
  CatVTON~\cite{chong2024catvton}  &  \drawgreen{\Checkmark}  & \drawred{\XSolidBrush}   & \drawred{\XSolidBrush}   & \drawred{\XSolidBrush}   \\
  \hline 
  Any2AnyTryon (Ours) & \drawgreen{\Checkmark}   & \drawgreen{\Checkmark}   & \drawgreen{\Checkmark}   & \drawgreen{\Checkmark}   \\
  
  \bottomrule
\end{tabular}
\vspace{0.2cm}
\end{table*}


\section{Introduction}

Inspired by the breakthroughs brought by Diffusion Models (DMs), the generation of human-related content has seen rapid development, with Virtual Try-On (VTON) being one of the most highly focused applications. Virtual Try-On aims to generate high-fidelity outfitted model images. Specifically, given a model image and a garment image, VTON methods are required to "dress" the model in the specified garment. This functionality has important applications in areas like online virtual try-ons and game design, making it significant for industries such as multimedia games. Previous methods, CatVTON\cite{chong2024catvton} and MMTryon\cite{zhang2024mmtryon}, utilize mask obtained by segment model and text as edit guidance, respectively, to achieve rational virtual try-on results. However, these methods are limited by the availability of high-quality {garment, model} pairs and struggle in complex generation scenarios, such as with "in-the-wild" model images. As a result, they fail to generate high-quality and ideal results. To address this, Stable Garment \cite{wang2024stablegarment} and MMTryon \cite{zhang2024mmtryon} proposed using generative models to replace the model and background in images, thus expanding the dataset. These methods have successfully achieved higher-quality and more generalized virtual try-on generation with the enlarged dataset. However, existing try-on methods often can only perform specific try-on tasks and have strict limitations on the input conditions provided by users. Additionally, the garment styles in datasets of existing methods remain insufficiently diverse, making it difficult for existing VTON methods to produce high-quality outfitted model images for garments in complex scenes.

To solve these problems, we propose a new user-friendly mask-free VTON generation framework, Any2AnyTryon. As illustrated in Table~\ref{Tab:method_func_compare}, existing VTON methods struggle to support multiple try-on generation tasks. In contrast, Any2AnyTryon can simultaneously fulfill multiple tasks based on user instructions. Firstly, we collect and integrate a large and diverse garment-model pair dataset, LAION-Garment, which includes garment and model images and corresponding user textual editing instructions. The outfitted images of the models include both "in-the-wild" and "in-the-shop" scenarios. Using this extensive dataset, we train a model capable of generating desired outfitted model images and garment images based on user editing instructions in both shop and wild settings. Because the dataset contains a rich set of training samples, the model can generate desired outputs based on user-provided editing instructions, making it more user-friendly compared to previous methods.

The architecture of our model ensures that all conditions are provided in a clean latent format within the same representation space as the target latent, as demonstrated in Fig.~\ref{fig:teaser}. This design allows the model to handle a variety of clothing tasks effectively. Additionally, we develop Adaptive Position Embedding, which adjusts the position embedding based on input textual prompts and image conditions, enabling Any2AnyTryon to generate high-quality outfitted model images and garment images simultaneously. Therefore, our Any2AnyTryon not only satisfies the ability to perform different try-on generation tasks within the same framework but also ensures high-quality generation results. To validate the generative capabilities of our method, we conduct experiments and compare our results with existing state-of-the-art methods. The results show that our approach generates outfitted images with better quality, finer details, and a more realistic appearance.

In summary, the main contributions of our Any2AnyTryon are as follows:
\vspace{-3mm}

\begin{itemize}




    \item We propose Any2AnyTryon based on DiT, modeling virtual try-on as a conditional generation task across multiple images. This unified framework integrates three tasks: virtual try-on, model and garment generation.
    \item We design a new model architecture, where all conditions share the same representation space with the output images so that our method can achieve high-fidelity VTON generation in diverse scenarios. Additionally, we introduce Adaptive Position Embedding, which adjusts the position embeddings based on input textual and image conditions. This enhancement allows Any2AnyTryon to seamlessly generate high-quality results for various VTON tasks using flexible and non-fixed conditional inputs.
    \item We collected a large-scale garment-model dataset, LAION-Garment, providing sufficient data for training VTON models to achieve high-quality outfitted model images. Extensive experiments and evaluations demonstrate that Any2AnyTryon outperforms other baseline methods in performance.

\end{itemize}

\section{Related Work}

\subsection{Diffusion Transformer} 
Recent advancements, particularly the introduction of latent diffusion models \cite{song2020denoising, rombach2022high, lu2022dpm, luo2023latent}, have significantly improved both the quality and efficiency of generative tasks \cite{zeng2023face, li2024zone, li2024uv, zeng2024controllable, fast, processpainter}. To further elevate generative capabilities, large-scale transformer architectures have been incorporated into these frameworks, resulting in cutting-edge models such as DiT \cite{peebles2023scalable}. Building on these foundational innovations, FLUX.1\cite{flux} is a powerful flow-based\cite{lipman2022flow} and DiT-based\cite{peebles2023scalable} model. FLUX.1 is critically acclaimed for its excellent prompt understanding, text rendering, and natural image generation. Prompt understanding capability can be attributed to the T5 text encoder \cite{raffel2020exploring} which is a powerful LLM. Many plugins such as FLUX.1-Fill inpainting and Flux.1-Redux variation model are developed based on the FLUX.1. IC-LoRA\cite{lhhuang2024iclora} is a notable technique to achieve personalization. OminiControl\cite{tan2024ominicontrol} design a unified adapter for DiT to enable subject-driven and spatially-aligned conditional generation.

\subsection{Reference-based Image Generation.} 

Reference-based Image Generation refers to generating customized outputs using images as conditions. Image Prompt Adaption methods~\cite{ye2023ip, zhang2024ssr, wang2024instantid, zeng2023ipdreamer, makeup, hair, anti} leverage adapter structures and ControlNet~\cite{zhang2023adding} to achieve generation consistency in appearance, character ID, and style, respectively. ReferenceNet~\cite{hu2024animate} is commonly employed in customization tasks, such as virtual try-on \cite{xu2024ootdiffusion, kim2024stableviton}, makeup transfer, and hairstyle transfer. It extracts features from reference images using a U-Net and injects them into the self-attention layers of the denoising network to ensure consistency in customization generation. However, the aforementioned methods are limited to specific conditions and single-task scenarios. Generating images based on complex textual prompts and a variable number of input images remains a significant challenge in image generation. Our Any2AnyTryon framework addresses this by utilizing intricate textual instructions and diverse model and garment images as input conditions, enabling high-quality and user-friendly Virtual Try-On (VTON) generation.

\subsection{Virtual Try-On.} 
The virtual try-on task has received increasing interest since the release of the VITON-HD~\cite{choi2021viton} dataset. With the development of generation methods, many works have achieved impressive generation results on the virtual try-on task. Earlier Tryon works~\cite{han2018viton} follow the warping and aggregation pipeline, where warping relies on Thin Plate Spline (TPS)~\cite{bookstein1989principal} or flow-based approaches~\cite{xie2023gp} and aggregation is commonly based on generative models such as GAN, diffusion, etc. Try-on has gained great improvement\cite{chong2024catvton, zhang2024mmtryon} based on a powerful text-to-image model. However, most methods still suffer from complex data pre-process due to the reliance on garment masks, model poses, and parsing results. Our Any2AnyTryon framework eliminates the dependence on masks, poses, or any other such conditions. Instead, it requires only user-provided textual instructions along with input model and garment images to generate the desired results, significantly enhancing convenience and usability.






\begin{figure*}
    \centering
    \includegraphics[width=0.9\linewidth]{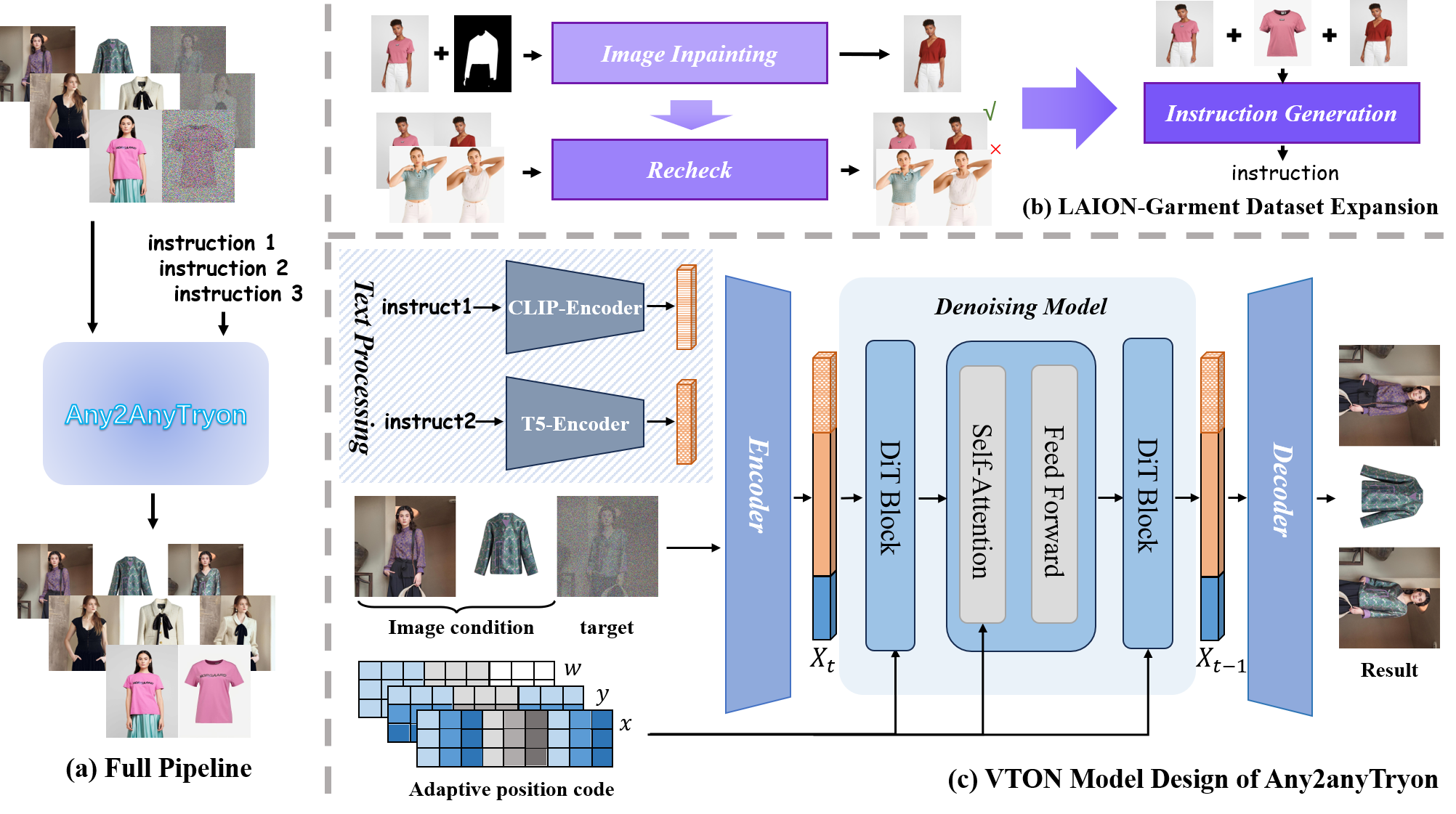}
    \caption{Overview of Any2AnyTryon.}
    \label{fig:framework}
\end{figure*}

\section{Method}

\subsection{Preliminaries}

\paragraph{Virtual Try-On Methods.}
Virtual Try-on methods can be broadly categorized into two types: Masked Try-On and Mask-free Try-On. Masked Try-On methods use segmentation masks to isolate regions of interest, such as clothing and the human body, allowing precise garment replacement or modification. The model $G$ generates the outfitted image $\hat{I}$ by applying the mask $M$ to the input model image $I$ and garment image $I_{gar}$, formulated as: 
\begin{equation}
\begin{aligned}
\hat{I} &= G(I, I_{gar}, M) \\
\end{aligned}
\end{equation}

This way ensures fine-grained control over the garment’s fit, as the mask restricts the clothing to the body region. On the other hand, Mask-free Try-On methods generate the try-on image $\hat{I}$ based on the model image $I$ and garment image $I_{gar}$, without relying on segmentation masks. This simplifies the process but often struggles with generalization and image quality. To overcome these limitations, we propose the mask-free VTON generation method Any2AnyTryon, which enhances both generalization and VTON quality.


\paragraph{Flow Matching}  
Flow matching~\cite{lipman2022flow} aligns the flow of information between noise and data distributions. It optimizes a velocity field that transforms noise into data over time, ensuring the generative model learns to map from the noise distribution to the true data distribution in a structured manner. The flow matching loss is formulated as follows:
\begin{equation}
L_{CFM} = E_{t, p_t(z|\epsilon), p(\epsilon)} \left[ \left\| v_\Theta(z, t) - u_t(z|\epsilon) \right\|^2 \right]
\end{equation}

Where \( v_\Theta(z, t) \) represents the velocity field parameterized by the neural network's weights, \( u_t(z|\epsilon) \) is the conditional vector field generated by the model to map the probabilistic path between the noise and true data distributions, and \( E \) denotes the expectation, involving integration or summation over time \( t \), conditional \( z \), and noise \( \epsilon \). This expectation calculates the average of the squared differences across all conditions, ensuring that the model's performance is averaged over many instances to provide a reliable measure of its generative capability.

\paragraph{FLUX.1}  
FLUX.1\cite{flux} is a powerful text-to-image model based on Flow Matching\cite{lipman2022flow} and DiT\cite{dehghani2023scaling}. It introduces rotary positional embeddings (RoPE)\cite{su2024roformer} and parallel attention layers\cite{dehghani2023scaling} to enhance performance and efficiency. FLUX.1 implements a three-dimensional RoPE scheme that supports three coordinates, but only the second and third dimensions are used for encoding spatial positions in the latent space:
\begin{equation}
\omega_m = \frac{1}{\theta^{2m/d}}, m \in [0, d/2)
\end{equation}
where $\theta$ is typically set to 10000. The position encoding applies a rotation matrix:
\begin{equation}
\begin{bmatrix} 
\cos(\omega_m \cdot \mathbf{pos}) & -\sin(\omega_m \cdot \mathbf{pos}) \\
\sin(\omega_m \cdot \mathbf{pos}) & \cos(\omega_m \cdot \mathbf{pos})
\end{bmatrix}
\end{equation}
This rotation is applied to query and key vectors in the attention mechanism, enabling the model to capture relative positional relationships in the latent space. Our Any2AnyTryon adjusts position embeddings to ensure more accurate garment fitting during the virtual try-on process.

\subsection{Overview of Any2anyTryon Framework}

In this work, we propose a new mask-free VTON generation method Any2AnyTryon, as demonstrated in Fig.~\ref{fig:framework}(a). Firstly, we collect and curate a new VTON dataset called LAION-Garment, which contains high-quality data and corresponding annotations for various VTON tasks. Meanwhile, we design a new VTON generation model that can support inputs of outfitted models and garment images with variable numbers and resolutions, and generate outputs based on user instructions, achieving user-friendly, controllable, and high-quality VTON generation. Next, we will detail how we collect the dataset in Section.~\ref{sec:method_data_collect}, and develop the Any2AnyTryon generation model in Section.~\ref{sec:method_model_design}.

\subsection{LAION-Garment Dataset Collection}
\label{sec:method_data_collect}


We use four public datasets including VITON-HD \cite{choi2021viton}, DressCode \cite{morelli2022dress}, DeepFashion2 \cite{DeepFashion2}, and LRVS-Fashion \cite{lepage2023lrvsf}. Most data are derived from the aforementioned datasets with the aid of auxiliary tools. Additionally, we manually crawl extra image pairs from the internet and filter the data to maintain quality.

To enhance Any2AnyTryon's generative capabilities across a broader range of VTON tasks, we further expand our LAION-Garment Dataset by integrating existing VTON datasets, as illustrated in Fig.~\ref{fig:framework}(b). First, to enrich the diversity of high-quality data, we augment the outfitted model–garment image pairs by incorporating images from both existing datasets and those crawled from the internet. Specifically, we utilize the Automasker from CatVTON \cite{chong2024catvton} and segment model \cite{guler2018densepose, ravi2024sam} to generate masks $M$ for the upper and lower garments in the outfitted model images and employed FLUX-Controlnet-Inpainting \cite{alibaba2024flux} to perform image inpainting on these masked images. This approach enables us to create additional data triples, each consisting of an outfitted model image $I$, a garment image $I_{gar}$, and the corresponding generated input model image $I_{\text{inpainted}}$. The image inpainting process can be represented as:
\begin{equation}
I_{\text{inpainted}} = \text{Inpainting}(I, M)
\end{equation}

To ensure the authenticity and quality of the generated images, we leverage GPT-4o \cite{achiam2023gpt} to select triples from the obtained data that exhibit strong authenticity in the generated model images, high quality, and strong consistency with the posture of the outfitted model images. These selected triples are then added to our LAION-Garment Dataset, thereby increasing the dataset's volume while maintaining its overall quality. Furthermore, to ensure that our LAION-Garment Dataset can encompass and support a wider range of VTON tasks, we utilize the Florence2 \cite{xiao2024florence} to generate user instructions for different tasks. Specifically, we reorganize and reclassify our curated LAION-Garment Dataset according to various tasks, then input the image data and description templates for each task into the Florence2 \cite{xiao2024florence}. This enables the generation of user instructions $T$ tailored for training the Any2AnyTryon model. For better understanding, we provide an example for the Virtual Try-On task: the instruction template is “The set of three images display a model, a garment, and the model wearing the garment. \texttt{<IMAGE1>} shows a person wearing the garment. \texttt{<IMAGE2>} depicts the garment. \texttt{<IMAGE3>} illustrates \texttt{<IMAGE1>} with \texttt{<IMAGE2>} worn by the model.”

\subsection{Any2AnyTryon Model Design}
\label{sec:method_model_design}

To enable our Any2AnyTryon to handle different VTON tasks within a unified module effectively, we concatenate image conditions and target noisy images along the pixel dimensions, as shown in Fig.~\ref{fig:framework}(c). This approach allows us to input model and garment images of varying quantities and resolutions, which together with user textual instructions, serve as conditions to guide VTON generation. Additionally, to ensure the high quality of the generated results, we train LoRA \cite{hu2021lora} based on the FLUX.1 model \cite{flux}.
We concatenate the denoised images and condition images on the pixel dimensions, which is equivalent to concatenating noised tokens and condition tokens. Velocity $v$ predicted by $v_\Theta$ can be rewritten as:
\begin{equation}
v = v_\Theta(\text{concat}[X; C_I; C_T], t),
\end{equation}
where $\text{concat}[X; C_I; C_T]$ denotes the concatenation of noised latent, tokens of image conditions $I$ and tokens of user instruction $T$.


Due to the concatenation of image conditions with variable quantities and resolutions with the output image at the pixel level in Any2AnyTryon, it becomes imperative to locate the position of each image condition within the input image. For virtual try-on tasks, it is necessary to align the non-edited regions of the input image conditions with the generated outfitted model images at the pixel level, thereby ensuring consistency of the non-edited regions. To this end, we propose Adaptive Position Embedding. Specifically, our approach utilizes a three-channel position encoding \( E_{pos} \in \mathbb{R}^{H \times W \times 3} \), where the first channel serves as a binary mask to separate different regions, while the second and third channels encode spatial positions along the height and width dimensions, respectively. The following formula shows our adaptive position embedding strategy:




\begin{equation}
R_{\Theta, p}^d(q) = 
\begin{bmatrix}
q_1 \\
q_2 \\
q_3 \\
q_4 \\
\vdots \\
q_{d-1} \\
q_{d} \\
\end{bmatrix}
\odot 
\begin{bmatrix}
\cos(p \theta_1) \\
\cos(p \theta_1) \\
\cos(p \theta_2) \\
\cos(p \theta_2) \\
\vdots \\
\cos(p \theta_{d/2}) \\
\cos(p \theta_{d/2}) \\
\end{bmatrix}
+
\begin{bmatrix}
-q_2 \\
q_1 \\
-q_4 \\
q_3 \\
\vdots \\
-q_{d} \\
q_{d-1} \\
\end{bmatrix}
\odot 
\begin{bmatrix}
\sin(p \theta_1) \\
\sin(p \theta_1) \\
\sin(p \theta_2) \\
\sin(p \theta_2) \\
\vdots \\
\sin(p \theta_{d/2}) \\
\sin(p \theta_{d/2}) \\
\end{bmatrix}
\label{Eq:rope}
\end{equation}

\begin{equation}
\resizebox{0.9\linewidth}{!}{$%
\begin{split}
Pos(X, C_I) &= Position(\begin{bmatrix} C_I & X \end{bmatrix}) \\
    &= \begin{bmatrix} 
    (y_1,x_1) & (y_1, x_2) & \ldots  & (y_1, x_r) & (y_1,x'_1) & \ldots  & (y_1, x'_s) \\
    (y_2,x_1) & (y_2, x_2) & \ldots  & (y_2, x_r) & (y_2,x'_1) & \ldots  & (y_2, x'_s) \\
    \ldots & \ldots & \ldots & \ldots & \ldots & \ldots & \ldots \\
    (y_m,x_1) & (y_m, x_2) & \ldots  & (y_m, x_r) & (y_m,x'_1) & \ldots  & (y_m, x'_s)
    \end{bmatrix} \\
x'_k &= \begin{cases}  x_k, & \text{if pixel-aligned} \\ 
x_{k+r}, & \text{otherwise } \end{cases}
\end{split}$%
\label{Eq:wpos}
}
\end{equation}

\begin{equation}
\begin{split}
Pos(q)[w] &= \begin{cases}  i, & \text{if is image condition $i$} \\ 
0, & \text{otherwise } \end{cases} \\
Pos(q)[y;x] &= Pos(X, C_I) \\
E_{pos}(q) = \mathrm{concat}[R&(Pos(q)[w]); R(Pos(q)[y]);R(Pos(q)[x])]
\end{split}
\label{Eq:ape}
\end{equation}

where $w, y, x$ represents three parts of the input vector corresponding to different position encodings: image condition ID, height, and width, the most common input vector is $q$ and $k$ in the attention mechanism, $r$ and $s$ represents number of reference and target tokens respectively. This module enables precise spatial alignment between input image conditions and generated images. Additionally, we introduce clean condition latents as input instead of noisy latents used in IC-LoRA \cite{lhhuang2024iclora}, as we found that background regions tend to change undesirably with noisy conditions. By maintaining clean latents in masked regions specified by split masks, our approach achieves superior preservation of unchanged areas while enabling precise control over target modifications.

After introducing the input form of conditions in Any2AnyTryon and incorporating Adaptive Position Embedding, the conditional flow matching loss can be represented as:
\begin{equation}
\resizebox{0.9\linewidth}{!}{$
L_{CFM} = E_{t, p_t(z|\epsilon), p(\epsilon)} \left[ \left\| v_\Theta(\text{concat}[X; C_I; C_T], E_{pos}, t) - u_t(z|\epsilon) \right\|^2 \right]
$
}
\end{equation}

\begin{figure}[!t]
\centering
\includegraphics[width=0.8\columnwidth]{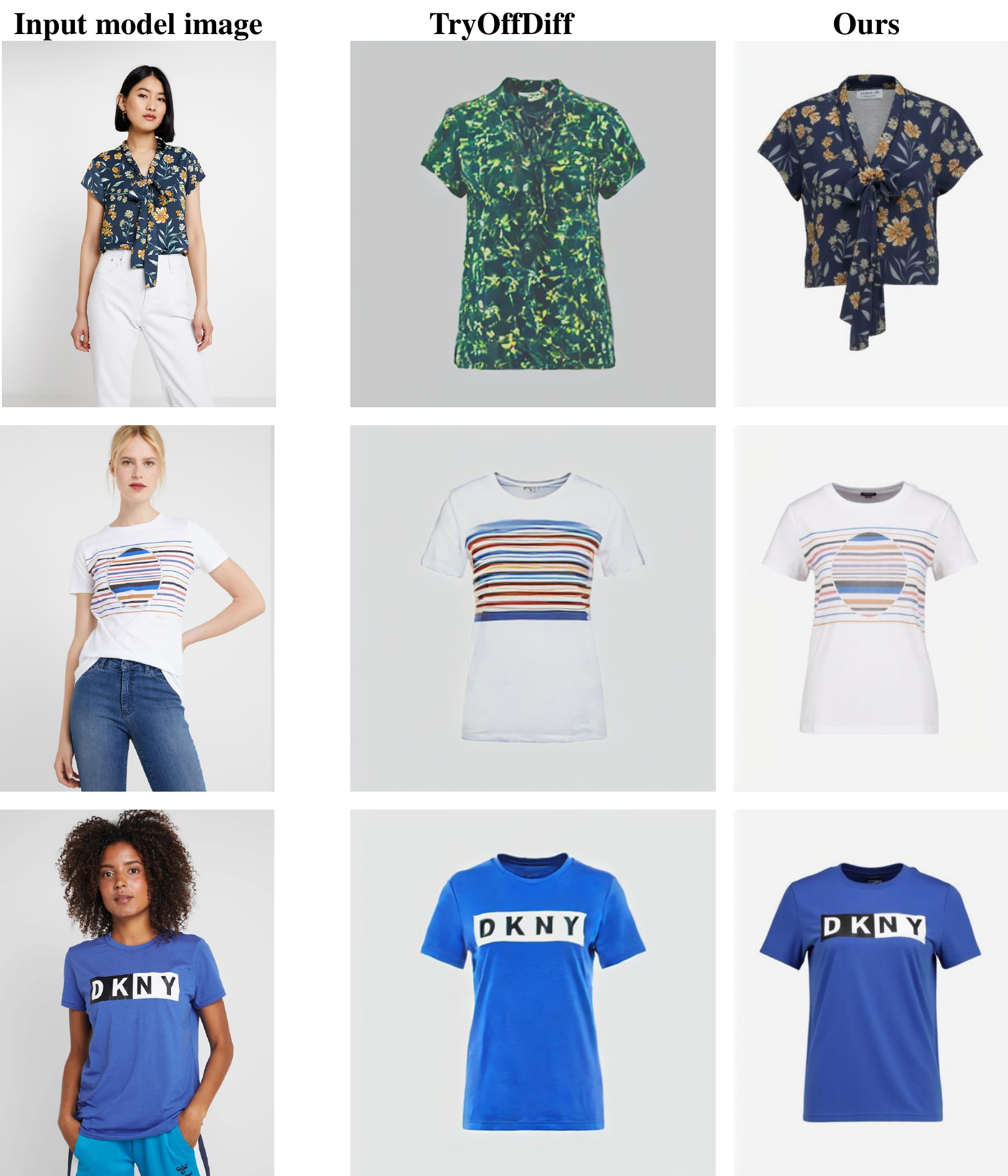}
\caption{Qualitative comparison of garment reconstruction.}
\label{Fig:exp_comp_garment}
\end{figure}

\begin{figure}[!t]
\centering
\includegraphics[width=0.8\columnwidth]{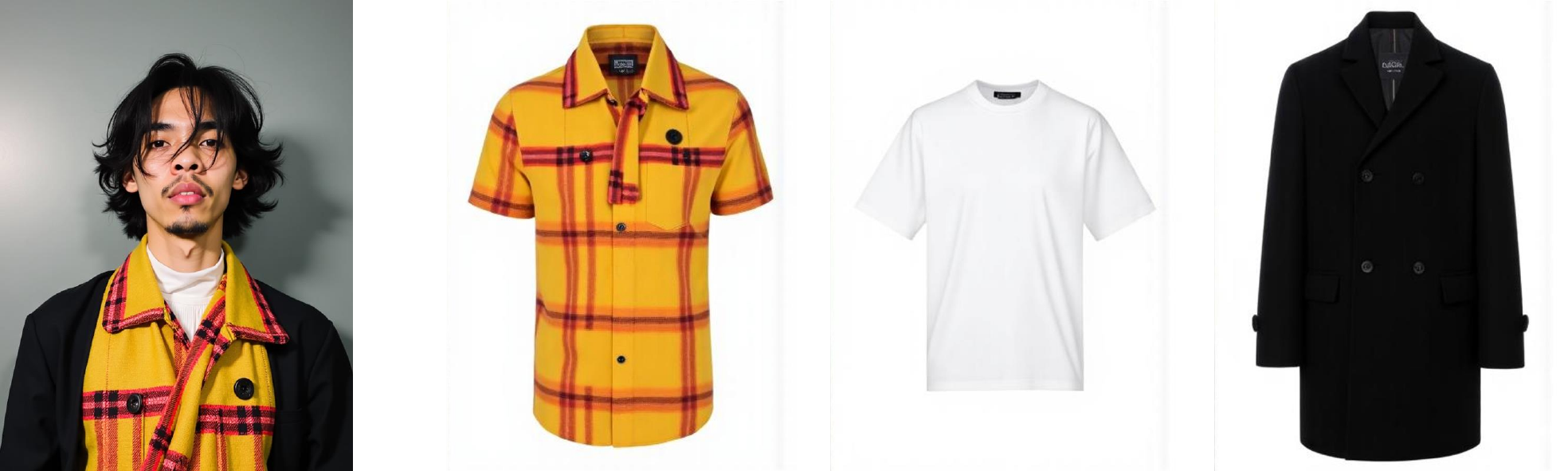}
\caption{Demonstration of multi garment flatten ability of our model with different prompt.}
\label{Fig:garment_flatten_quality}
\end{figure}

\begin{table*}[!ht]
  \small
  \centering
  \setlength{\abovecaptionskip}{0cm}
  \caption{Quantitative comparison on VITON-HD dataset.}
  \label{Tab:metrics_garment_flatten}
  \vspace{0.2cm}
\begin{tabular}{l|cccccccc}
  \toprule
  \textbf{Method} & \textbf{SSIM~\cite{wang2004ssim}}↑ & \textbf{MS-SSIM}↑ & \textbf{CW-SSIM}↑  & \textbf{LPIPS~\cite{zhang2018lpips}}↓  & \textbf{FID~\cite{heusel2017fid}}↓ & \textbf{CLIP-FID}↓ & \textbf{KID~\cite{binkowski2018kid}}↓ & \textbf{DISTS~\cite{ding2020dists}}↓  \\
  \midrule
  TryOffDiff~\cite{velioglu2024tryoffdiff}  & 0.793 & \textbf{0.712} & \textbf{0.466} & 0.334 & 20.346 & 8.371 & 6.8 & 0.226 \\
  Ours & \textbf{0.805} & 0.710 & 0.453 & \textbf{0.328} & \textbf{13.367} & \textbf{3.872} & \textbf{3.5} & \textbf{0.217} \\
  \bottomrule
\end{tabular}
\vspace{0.2cm}
\end{table*}

\begin{figure}[!t]
\centering
\includegraphics[width=\columnwidth]{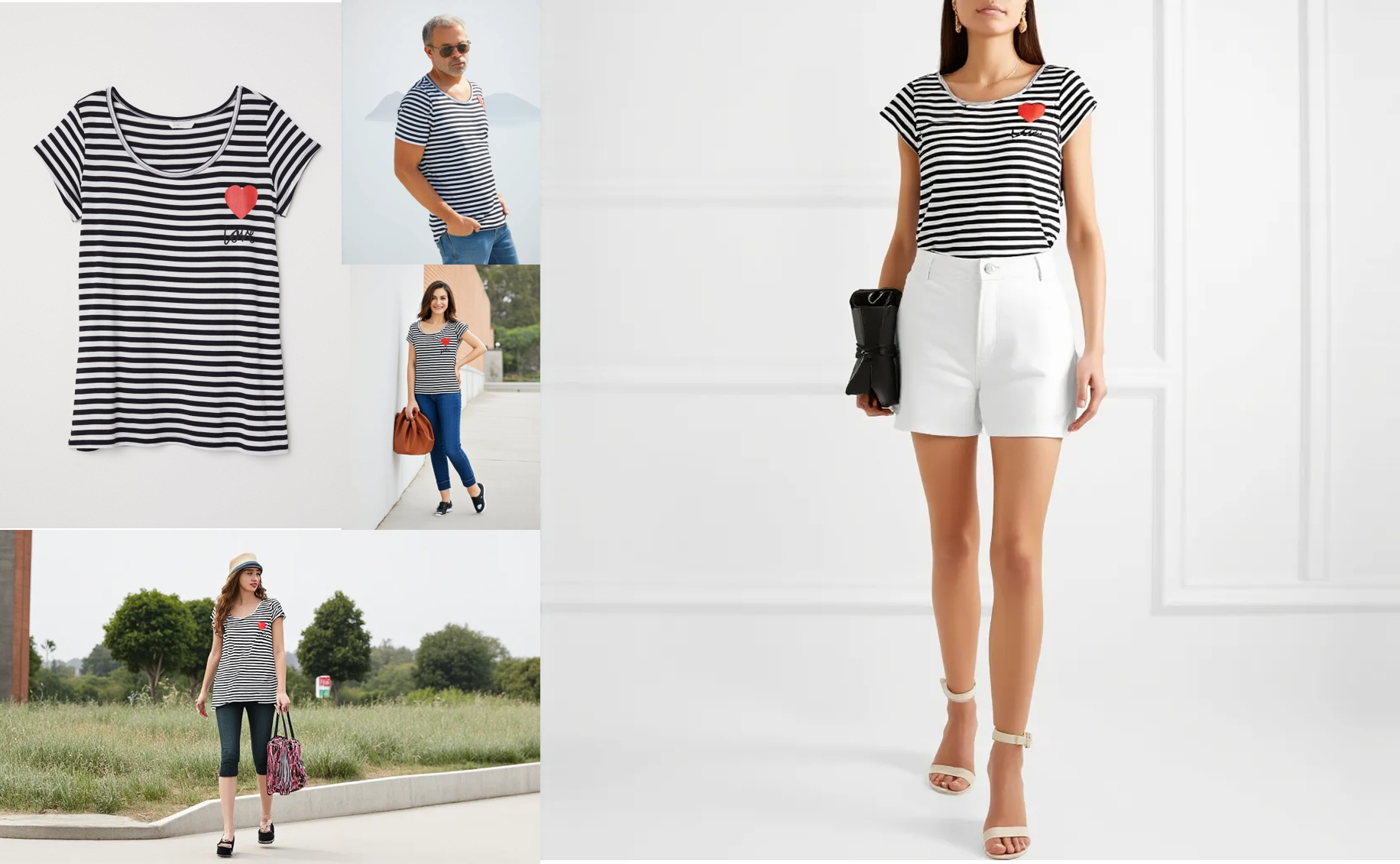}
\caption{Demonstration of garment-driven generation ability of our model with different image sizes.}
\label{Fig:model_generation_quality}
\end{figure}

\begin{table}[!ht]
  \small
  \centering
  \setlength{\abovecaptionskip}{0cm}
  \caption{Quantitative comparison on VITON-HD dataset.}
  \label{Tab:model_free_vton}
  \vspace{0.2cm}
  \resizebox{\columnwidth}{!}{
\begin{tabular}{l|ccccc}
  \toprule
  \textbf{Method} & \textbf{MP-LPIPS~\cite{chen2024magic}}↓ & \textbf{CLIP-I~\cite{radford2021learning}}↑ &\textbf{DiffSim~\cite{song2024diffsimtamingdiffusionmodels}}↑
  &\textbf{FFA~\cite{kotar2023these}}↑
  \\ 
  \midrule
  Magic Clothing~\cite{chen2024magic}    & 0.192 & 0.642 & 0.143 & 0.459 \\    
  StableGarment~\cite{wang2024stablegarment}    & 0.149 & 0.650 & 0.153 & 0.547 \\  
  Ours   &  \textbf{0.141}  & \textbf{0.789}  & \textbf{0.202} & \textbf{0.549} \\   
  \bottomrule
\end{tabular}
}
\vspace{0.2cm}
\end{table}

\begin{figure*}[!th]
\centering
\includegraphics[width=1.8\columnwidth]{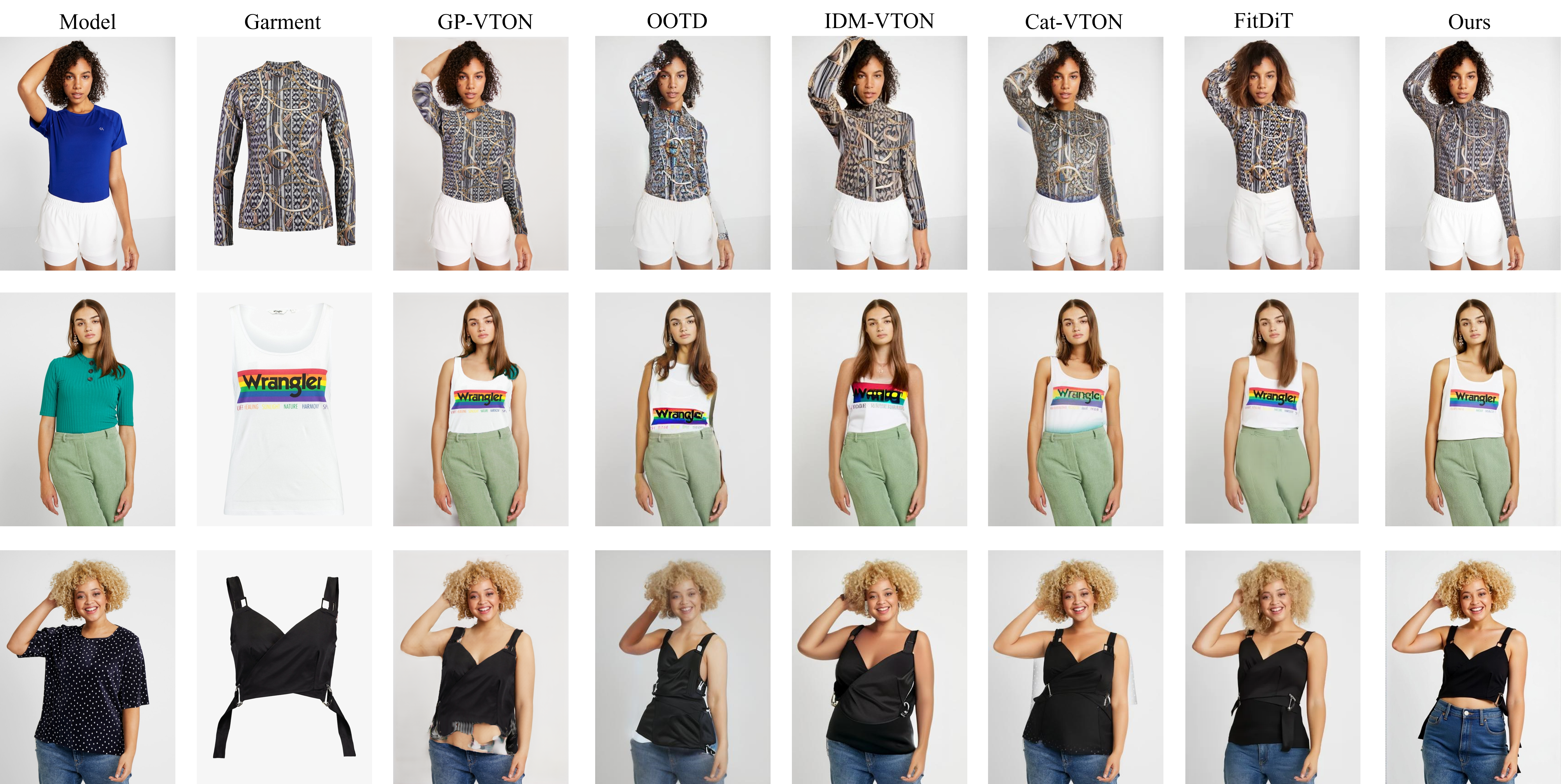}
\caption{Qualitative comparison of virtual try-on on the VITONHD dataset.}
\label{Fig:tryon_quality}
\end{figure*}

\begin{table}[!ht]
  \small
  \centering
  \setlength{\abovecaptionskip}{0cm}
  \caption{Quantitative comparison on VITON-HD dataset. We multiply KID by 1000 for better comparison. The best and the second best results are denoted as \textbf{Bold} and \underline{underline}, respectively.}
  \label{Tab:vton}
  \vspace{0.2cm}
\scalebox{0.95}{
\begin{tabular}{l|cccccc}
    \toprule
    \multirow{2}{*}{\textbf{Method}} & \multicolumn{4}{c}{Paired} & \multicolumn{2}{c}{Unpaired} \\
    \cmidrule(lr){2-5} \cmidrule(lr){6-7}
    & \textbf{LPIPS}↓ & \textbf{SSIM}↑ & \textbf{FID}↓ & \textbf{KID}↓ & \textbf{FID}↓ & \textbf{KID}↓ \\
  \midrule
  GP-VTON\cite{xie2023gp}   &  \underline{0.0677}  & \textbf{0.8722} & 8.649 & 3.669 & 11.708 & 3.990 \\
  OOTD\cite{xu2024ootdiffusion} & 0.1317 & 0.7838 & 12.131 & 4.335  & 15.136 & 5.774 \\
  IDM-VTON\cite{choi2024improving} & 0.0815 & 0.8156 & 8.206 & 1.727  & 10.745 & 2.229 \\
  CatVTON~\cite{chong2024catvton}    & \textbf{0.0582} & \underline{0.8653} & \textbf{5.482} & \textbf{0.384} & \underline{9.083} & \underline{1.130} \\
  FitDiT\cite{jiang2024fitdit}   &  0.1059  & 0.8298 & 8.362 & 1.543 & 10.340 & 1.648 \\
  Ours  &  0.0877 & 0.8387 & \underline{6.934} & \underline{0.7387} & \textbf{8.965} & \textbf{0.981} \\
  \bottomrule
\end{tabular}}
\vspace{0.2cm}
\end{table}

\begin{table}[!ht]
  \small
  \centering
  \setlength{\abovecaptionskip}{0cm}
  \caption{Ablation study on proposed components.}
  \label{Tab:ablation_pos}
  \vspace{0.2cm}
  \resizebox{\columnwidth}{!}{
\begin{tabular}{l|ccccc}
  \toprule
  \textbf{Setting} & \textbf{LPIPS}↓ & \textbf{SSIM}↑ &\textbf{CLIP-FID}↓ &\textbf{KID}↓
  \\
  \midrule
  Ours(w/o clean latent) & 0.3141 & 0.6892 & 9.648 & 6.403 \\
  Ours(w/o adaptive position)  & 0.3080 & 0.7088 & 9.434 & 5.658 \\
  Ours(full) & \textbf{0.2590} & \textbf{0.7373} & \textbf{9.293} & \textbf{5.407} \\
  \bottomrule
\end{tabular}
}
\vspace{0.2cm}
\end{table}

\begin{figure}[!t]
\centering
\includegraphics[width=0.8\columnwidth]{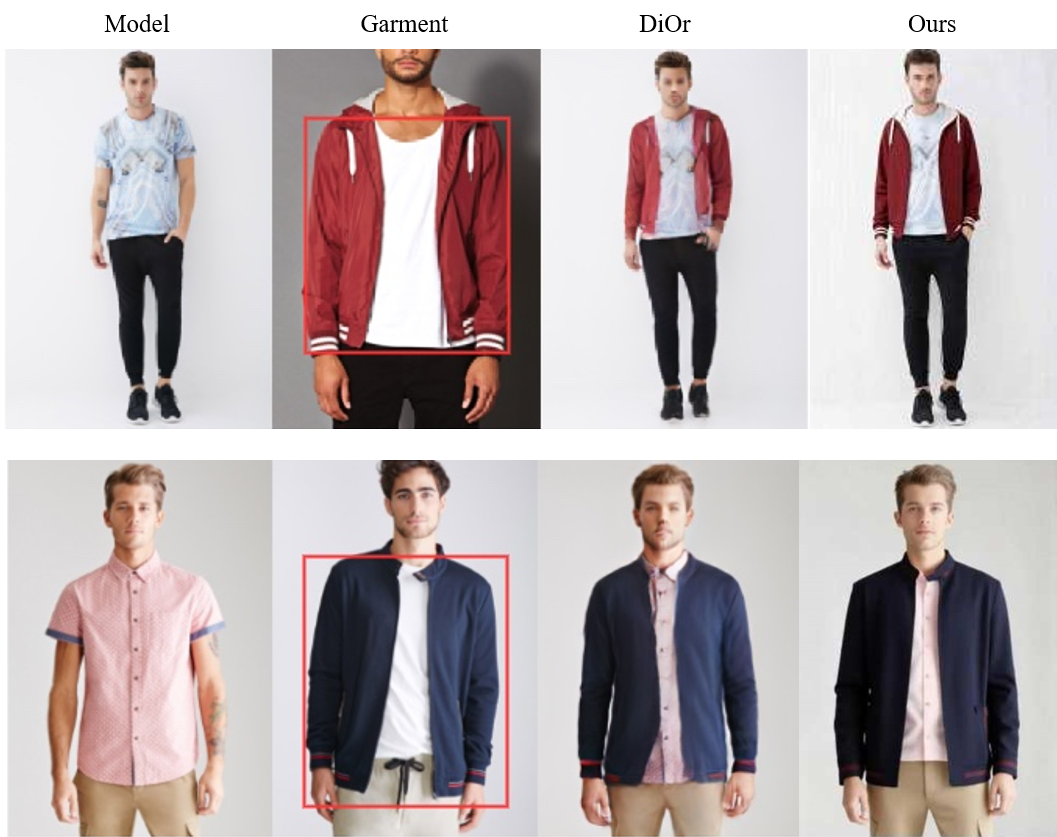}
\caption{Demonstration of try-on in layer.}
\label{Fig:tryon_layer}
\end{figure}

\section{Experiment}

\subsection{Implementaion Details}
Our method is based on the dev version of FLUX.1 model\cite{flux}. The garment reconstruction model, model-free virtual try-on model and try-on model for evaluation are trained with height and width set as 512x384, 768x576 and 512x384, respectively. All the figures presented in the paper are generated from the unified model using all the data with all the tasks in variable image size. To train the unified model, we train all the tasks except try-on in layers in the first stage and finetune with the dataset for try-on in layers and subset from the rest of the tasks in the second stage. In all training, we use prodigy optimizer\cite{mishchenko2023prodigy} with weight decay set as 0.01. 

For the LAION-Garment dataset, we include several existing VTON datasets for the try-on training task, which consist of the following: VITON-HD \cite{choi2021viton} with 11,491 image pairs; DressCode \cite{morelli2022dress} with 13,563 image pairs of Upper body, 27,678 image pairs of Dresses, and 8,689 image pairs of Lower body; DeepFashion2 \cite{DeepFashion2} with 988 image pairs. After data augmentation and user instructions generation, the dataset expands to over 60,000 data triples. The validation metrics are detailed in the supplementary material.

\subsection{Garment Reconstruction}
Garment reconstruction refers to reconstructing flattened garments from images of the model wearing the target garment. We support multi-garment reconstruction from our model with different prompts as shown in Fig.~\ref{Fig:garment_flatten_quality}. To validate the effectiveness of our methods, we compare our results with TryOffDiff\cite{velioglu2024tryoffdiff}. As shown in Table.~\ref{Tab:metrics_garment_flatten}, Compared to existing methods, the garments reconstructed by our approach not only have higher quality but also match the appearance of the garment worn by the model in the input image more accurately. To more intuitively demonstrate the effectiveness of our Any2AnyTryon garment reconstruction, as shown in Fig.~\ref{Fig:exp_comp_garment}, it is evident that the realism and accuracy of garment reconstruction achieved by our method are far superior to those of TryOffDiff, further validating the effectiveness of our approach.

\subsection{Model-free Virtual Try-on}
Model-free Virtual Try-on refers to garment-driven model generation which can be regarded as a specific sub task of subject-driven generation. We compared with Magic Clothing~\cite{chen2024magic} and StableGarment~\cite{wang2024stablegarment}. In Table~\ref{Tab:model_free_vton}, we compare our Any2AnyTryon and baseline methods on the Model-free VTON generation task using the VITON-HD dataset. We employed five metrics, including DiffSim~\cite{song2024diffsimtamingdiffusionmodels} and FFA~\cite{kotar2023these}, for a quantitative comparison. The results indicate that our Any2AnyTryon significantly outperforms baseline methods to preserve the original garment appearance and generate high-quality outfitted model images. Additionally, our Any2AnyTryon can achieve high-quality VTON generation at various image resolutions, demonstrated in Fig. \ref{Fig:model_generation_quality}.

\subsection{Virtual Try-on}
We compare our mask-free try-on results with GP-VTON\cite{xie2023gp}, OOTD\cite{xu2024ootdiffusion}, IDM-VTON\cite{choi2024improving}, CatVTON\cite{chong2024catvton} and FitDiT\cite{jiang2024fitdit}. As shown in Table~\ref{Tab:vton}, overall, our method outperforms existing state-of-the-art (SOTA) virtual try-on generation methods, particularly in terms of the FID~\cite{heusel2017fid} and KID~\cite{binkowski2018kid} metrics, which measure generation quality. Additionally, for virtual try-on generation, the visualization of outfitted model images is crucial. In Fig.~\ref{Fig:tryon_quality}, we present the generation results of different methods. It is evident that our Any2AnyTryon not only produces higher-quality and more realistic outfitted model images but also ensures that the garments worn by the model align closely with the input garments. This demonstrates the superior generation capability of our Any2AnyTryon in virtual try-on tasks.

\subsection{Try-on in layers}
We compare with DiOr\cite{cui2021dressing} which proposes similar try-on-in-layer tasks, also called dressing in order there. The try-on-in-layer task is very challenging, especially for mask-free try-on, as the method must identify specific editing locations only with text guidance. To showcase the generation capability of our Any2AnyTryon, we performed a qualitative comparison using examples from DiOr's original paper. As depicted in Fig.~\ref{Fig:tryon_layer}, our method better preserves the appearance of the input model, while also aligning the garment worn by the outfitted model with the input garment more accurately.

\subsection{Ablation Study}
To further demonstrate the effectiveness of the image condition addition strategy and Adaptive Position Embedding in our Any2AnyTryon, we conduct an ablation study. In Table~\ref{Tab:ablation_pos}, we compare two different methods. One method differs from the image condition addition in Any2AnyTryon by adding noise to the image condition and generation target as input to the model. The second method is without the adaptive position embedding. The results show that, compared to the full Any2AnyTryon method, both comparison methods negatively impact the generation results, thereby confirming the validity of the Any2AnyTryon design.

\section{Conclusion}
The proposed Any2anyTryon framework demonstrates significant advancements in virtual try-on tasks by introducing a unified, mask-free solution capable of handling diverse scenarios. By leveraging the LAION-Garment dataset and innovative techniques such as adaptive position embedding and clean condition latents, the method enhances garment reconstruction, model-free virtual try-on, and layered try-on tasks. Extensive experiments validate its effectiveness, showcasing superior performance in generating high-fidelity, realistic outfitted images compared to state-of-the-art methods. The framework’s flexibility, scalability, and ability to generalize across complex conditions mark a significant step forward in virtual try-on research and its practical applications.

\begin{acks}
To Robert, for the bagels and explaining CMYK and color spaces.
\end{acks}

\bibliographystyle{ACM-Reference-Format}
\bibliography{main}


\begin{thebibliography}{59}


\ifx \showCODEN    \undefined \def \showCODEN     #1{\unskip}     \fi
\ifx \showDOI      \undefined \def \showDOI       #1{#1}\fi
\ifx \showISBNx    \undefined \def \showISBNx     #1{\unskip}     \fi
\ifx \showISBNxiii \undefined \def \showISBNxiii  #1{\unskip}     \fi
\ifx \showISSN     \undefined \def \showISSN      #1{\unskip}     \fi
\ifx \showLCCN     \undefined \def \showLCCN      #1{\unskip}     \fi
\ifx \shownote     \undefined \def \shownote      #1{#1}          \fi
\ifx \showarticletitle \undefined \def \showarticletitle #1{#1}   \fi
\ifx \showURL      \undefined \def \showURL       {\relax}        \fi
\providecommand\bibfield[2]{#2}
\providecommand\bibinfo[2]{#2}
\providecommand\natexlab[1]{#1}
\providecommand\showeprint[2][]{arXiv:#2}

\bibitem[Achiam et~al\mbox{.}(2023)]%
        {achiam2023gpt}
\bibfield{author}{\bibinfo{person}{Josh Achiam}, \bibinfo{person}{Steven Adler}, \bibinfo{person}{Sandhini Agarwal}, \bibinfo{person}{Lama Ahmad}, \bibinfo{person}{Ilge Akkaya}, \bibinfo{person}{Florencia~Leoni Aleman}, \bibinfo{person}{Diogo Almeida}, \bibinfo{person}{Janko Altenschmidt}, \bibinfo{person}{Sam Altman}, \bibinfo{person}{Shyamal Anadkat}, {et~al\mbox{.}}} \bibinfo{year}{2023}\natexlab{}.
\newblock \showarticletitle{Gpt-4 technical report}.
\newblock \bibinfo{journal}{\emph{arXiv preprint arXiv:2303.08774}} (\bibinfo{year}{2023}).
\newblock


\bibitem[Alibaba(2024)]%
        {alibaba2024flux}
\bibfield{author}{\bibinfo{person}{Alibaba}.} \bibinfo{year}{2024}\natexlab{}.
\newblock \bibinfo{title}{FLUX-Controlnet-Inpainting}.
\newblock
\newblock
\urldef\tempurl%
\url{https://github.com/alimama-creative/FLUX-Controlnet-Inpainting}
\showURL{%
\tempurl}


\bibitem[Bi{\'n}kowski et~al\mbox{.}(2018)]%
        {binkowski2018kid}
\bibfield{author}{\bibinfo{person}{Miko{\l}aj Bi{\'n}kowski}, \bibinfo{person}{Danica~J Sutherland}, \bibinfo{person}{Michael Arbel}, {and} \bibinfo{person}{Arthur Gretton}.} \bibinfo{year}{2018}\natexlab{}.
\newblock \showarticletitle{Demystifying mmd gans}.
\newblock \bibinfo{journal}{\emph{arXiv preprint arXiv:1801.01401}} (\bibinfo{year}{2018}).
\newblock


\bibitem[Bookstein(1989)]%
        {bookstein1989principal}
\bibfield{author}{\bibinfo{person}{Fred~L. Bookstein}.} \bibinfo{year}{1989}\natexlab{}.
\newblock \showarticletitle{Principal warps: Thin-plate splines and the decomposition of deformations}.
\newblock \bibinfo{journal}{\emph{IEEE Transactions on pattern analysis and machine intelligence}} \bibinfo{volume}{11}, \bibinfo{number}{6} (\bibinfo{year}{1989}), \bibinfo{pages}{567--585}.
\newblock


\bibitem[Chen et~al\mbox{.}(2024)]%
        {chen2024magic}
\bibfield{author}{\bibinfo{person}{Weifeng Chen}, \bibinfo{person}{Tao Gu}, \bibinfo{person}{Yuhao Xu}, {and} \bibinfo{person}{Arlene Chen}.} \bibinfo{year}{2024}\natexlab{}.
\newblock \showarticletitle{Magic clothing: Controllable garment-driven image synthesis}. In \bibinfo{booktitle}{\emph{Proceedings of the 32nd ACM International Conference on Multimedia}}. \bibinfo{pages}{6939--6948}.
\newblock


\bibitem[Choi et~al\mbox{.}(2021)]%
        {choi2021viton}
\bibfield{author}{\bibinfo{person}{Seunghwan Choi}, \bibinfo{person}{Sunghyun Park}, \bibinfo{person}{Minsoo Lee}, {and} \bibinfo{person}{Jaegul Choo}.} \bibinfo{year}{2021}\natexlab{}.
\newblock \showarticletitle{Viton-hd: High-resolution virtual try-on via misalignment-aware normalization}. In \bibinfo{booktitle}{\emph{Proceedings of the IEEE/CVF conference on computer vision and pattern recognition}}. \bibinfo{pages}{14131--14140}.
\newblock


\bibitem[Choi et~al\mbox{.}(2024)]%
        {choi2024improving}
\bibfield{author}{\bibinfo{person}{Yisol Choi}, \bibinfo{person}{Sangkyung Kwak}, \bibinfo{person}{Kyungmin Lee}, \bibinfo{person}{Hyungwon Choi}, {and} \bibinfo{person}{Jinwoo Shin}.} \bibinfo{year}{2024}\natexlab{}.
\newblock \showarticletitle{Improving Diffusion Models for Authentic Virtual Try-on in the Wild}.
\newblock \bibinfo{journal}{\emph{arXiv preprint arXiv:2403.05139}} (\bibinfo{year}{2024}).
\newblock


\bibitem[Chong et~al\mbox{.}(2024)]%
        {chong2024catvton}
\bibfield{author}{\bibinfo{person}{Zheng Chong}, \bibinfo{person}{Xiao Dong}, \bibinfo{person}{Haoxiang Li}, \bibinfo{person}{Shiyue Zhang}, \bibinfo{person}{Wenqing Zhang}, \bibinfo{person}{Xujie Zhang}, \bibinfo{person}{Hanqing Zhao}, {and} \bibinfo{person}{Xiaodan Liang}.} \bibinfo{year}{2024}\natexlab{}.
\newblock \showarticletitle{CatVTON: Concatenation Is All You Need for Virtual Try-On with Diffusion Models}.
\newblock \bibinfo{journal}{\emph{arXiv preprint arXiv:2407.15886}} (\bibinfo{year}{2024}).
\newblock


\bibitem[Cui et~al\mbox{.}(2021)]%
        {cui2021dressing}
\bibfield{author}{\bibinfo{person}{Aiyu Cui}, \bibinfo{person}{Daniel McKee}, {and} \bibinfo{person}{Svetlana Lazebnik}.} \bibinfo{year}{2021}\natexlab{}.
\newblock \showarticletitle{Dressing in order: Recurrent person image generation for pose transfer, virtual try-on and outfit editing}. In \bibinfo{booktitle}{\emph{Proceedings of the IEEE/CVF international conference on computer vision}}. \bibinfo{pages}{14638--14647}.
\newblock


\bibitem[Dehghani et~al\mbox{.}(2023)]%
        {dehghani2023scaling}
\bibfield{author}{\bibinfo{person}{Mostafa Dehghani}, \bibinfo{person}{Josip Djolonga}, \bibinfo{person}{Basil Mustafa}, \bibinfo{person}{Piotr Padlewski}, \bibinfo{person}{Jonathan Heek}, \bibinfo{person}{Justin Gilmer}, \bibinfo{person}{Andreas~Peter Steiner}, \bibinfo{person}{Mathilde Caron}, \bibinfo{person}{Robert Geirhos}, \bibinfo{person}{Ibrahim Alabdulmohsin}, {et~al\mbox{.}}} \bibinfo{year}{2023}\natexlab{}.
\newblock \showarticletitle{Scaling vision transformers to 22 billion parameters}. In \bibinfo{booktitle}{\emph{International Conference on Machine Learning}}. PMLR, \bibinfo{pages}{7480--7512}.
\newblock


\bibitem[Ding et~al\mbox{.}(2020)]%
        {ding2020dists}
\bibfield{author}{\bibinfo{person}{Keyan Ding}, \bibinfo{person}{Kede Ma}, \bibinfo{person}{Shiqi Wang}, {and} \bibinfo{person}{Eero~P Simoncelli}.} \bibinfo{year}{2020}\natexlab{}.
\newblock \showarticletitle{Image quality assessment: Unifying structure and texture similarity}.
\newblock \bibinfo{journal}{\emph{IEEE transactions on pattern analysis and machine intelligence}} \bibinfo{volume}{44}, \bibinfo{number}{5} (\bibinfo{year}{2020}), \bibinfo{pages}{2567--2581}.
\newblock


\bibitem[Ge et~al\mbox{.}(2019)]%
        {DeepFashion2}
\bibfield{author}{\bibinfo{person}{Yuying Ge}, \bibinfo{person}{Ruimao Zhang}, \bibinfo{person}{Lingyun Wu}, \bibinfo{person}{Xiaogang Wang}, \bibinfo{person}{Xiaoou Tang}, {and} \bibinfo{person}{Ping Luo}.} \bibinfo{year}{2019}\natexlab{}.
\newblock \showarticletitle{A Versatile Benchmark for Detection, Pose Estimation, Segmentation and Re-Identification of Clothing Images}.
\newblock \bibinfo{journal}{\emph{CVPR}} (\bibinfo{year}{2019}).
\newblock


\bibitem[G{\"u}ler et~al\mbox{.}(2018)]%
        {guler2018densepose}
\bibfield{author}{\bibinfo{person}{R{\i}za~Alp G{\"u}ler}, \bibinfo{person}{Natalia Neverova}, {and} \bibinfo{person}{Iasonas Kokkinos}.} \bibinfo{year}{2018}\natexlab{}.
\newblock \showarticletitle{Densepose: Dense human pose estimation in the wild}. In \bibinfo{booktitle}{\emph{Proceedings of the IEEE conference on computer vision and pattern recognition}}. \bibinfo{pages}{7297--7306}.
\newblock


\bibitem[Han et~al\mbox{.}(2018)]%
        {han2018viton}
\bibfield{author}{\bibinfo{person}{Xintong Han}, \bibinfo{person}{Zuxuan Wu}, \bibinfo{person}{Zhe Wu}, \bibinfo{person}{Ruichi Yu}, {and} \bibinfo{person}{Larry~S Davis}.} \bibinfo{year}{2018}\natexlab{}.
\newblock \showarticletitle{Viton: An image-based virtual try-on network}. In \bibinfo{booktitle}{\emph{Proceedings of the IEEE conference on computer vision and pattern recognition}}. \bibinfo{pages}{7543--7552}.
\newblock


\bibitem[Heusel et~al\mbox{.}(2017)]%
        {heusel2017fid}
\bibfield{author}{\bibinfo{person}{Martin Heusel}, \bibinfo{person}{Hubert Ramsauer}, \bibinfo{person}{Thomas Unterthiner}, \bibinfo{person}{Bernhard Nessler}, {and} \bibinfo{person}{Sepp Hochreiter}.} \bibinfo{year}{2017}\natexlab{}.
\newblock \showarticletitle{Gans trained by a two time-scale update rule converge to a local nash equilibrium}.
\newblock \bibinfo{journal}{\emph{Advances in neural information processing systems}}  \bibinfo{volume}{30} (\bibinfo{year}{2017}).
\newblock


\bibitem[Hu et~al\mbox{.}(2021)]%
        {hu2021lora}
\bibfield{author}{\bibinfo{person}{Edward~J Hu}, \bibinfo{person}{Yelong Shen}, \bibinfo{person}{Phillip Wallis}, \bibinfo{person}{Zeyuan Allen-Zhu}, \bibinfo{person}{Yuanzhi Li}, \bibinfo{person}{Shean Wang}, \bibinfo{person}{Lu Wang}, {and} \bibinfo{person}{Weizhu Chen}.} \bibinfo{year}{2021}\natexlab{}.
\newblock \showarticletitle{Lora: Low-rank adaptation of large language models}.
\newblock \bibinfo{journal}{\emph{arXiv preprint arXiv:2106.09685}} (\bibinfo{year}{2021}).
\newblock


\bibitem[Hu(2024)]%
        {hu2024animate}
\bibfield{author}{\bibinfo{person}{Li Hu}.} \bibinfo{year}{2024}\natexlab{}.
\newblock \showarticletitle{Animate anyone: Consistent and controllable image-to-video synthesis for character animation}. In \bibinfo{booktitle}{\emph{Proceedings of the IEEE/CVF Conference on Computer Vision and Pattern Recognition}}. \bibinfo{pages}{8153--8163}.
\newblock


\bibitem[Huang et~al\mbox{.}(2024)]%
        {lhhuang2024iclora}
\bibfield{author}{\bibinfo{person}{Lianghua Huang}, \bibinfo{person}{Wei Wang}, \bibinfo{person}{Zhi-Fan Wu}, \bibinfo{person}{Yupeng Shi}, \bibinfo{person}{Huanzhang Dou}, \bibinfo{person}{Chen Liang}, \bibinfo{person}{Yutong Feng}, \bibinfo{person}{Yu Liu}, {and} \bibinfo{person}{Jingren Zhou}.} \bibinfo{year}{2024}\natexlab{}.
\newblock \showarticletitle{In-Context LoRA for Diffusion Transformers}.
\newblock \bibinfo{journal}{\emph{arXiv preprint arxiv:2410.23775}} (\bibinfo{year}{2024}).
\newblock


\bibitem[Jiang et~al\mbox{.}(2024)]%
        {jiang2024fitdit}
\bibfield{author}{\bibinfo{person}{Boyuan Jiang}, \bibinfo{person}{Xiaobin Hu}, \bibinfo{person}{Donghao Luo}, \bibinfo{person}{Qingdong He}, \bibinfo{person}{Chengming Xu}, \bibinfo{person}{Jinlong Peng}, \bibinfo{person}{Jiangning Zhang}, \bibinfo{person}{Chengjie Wang}, \bibinfo{person}{Yunsheng Wu}, {and} \bibinfo{person}{Yanwei Fu}.} \bibinfo{year}{2024}\natexlab{}.
\newblock \showarticletitle{FitDiT: Advancing the Authentic Garment Details for High-fidelity Virtual Try-on}.
\newblock \bibinfo{journal}{\emph{arXiv preprint arXiv:2411.10499}} (\bibinfo{year}{2024}).
\newblock


\bibitem[Kim et~al\mbox{.}(2024)]%
        {kim2024stableviton}
\bibfield{author}{\bibinfo{person}{Jeongho Kim}, \bibinfo{person}{Guojung Gu}, \bibinfo{person}{Minho Park}, \bibinfo{person}{Sunghyun Park}, {and} \bibinfo{person}{Jaegul Choo}.} \bibinfo{year}{2024}\natexlab{}.
\newblock \showarticletitle{Stableviton: Learning semantic correspondence with latent diffusion model for virtual try-on}. In \bibinfo{booktitle}{\emph{Proceedings of the IEEE/CVF Conference on Computer Vision and Pattern Recognition}}. \bibinfo{pages}{8176--8185}.
\newblock


\bibitem[Kotar et~al\mbox{.}(2023)]%
        {kotar2023these}
\bibfield{author}{\bibinfo{person}{Klemen Kotar}, \bibinfo{person}{Stephen Tian}, \bibinfo{person}{Hong-Xing Yu}, \bibinfo{person}{Dan Yamins}, {and} \bibinfo{person}{Jiajun Wu}.} \bibinfo{year}{2023}\natexlab{}.
\newblock \showarticletitle{Are these the same apple? comparing images based on object intrinsics}.
\newblock \bibinfo{journal}{\emph{Advances in Neural Information Processing Systems}}  \bibinfo{volume}{36} (\bibinfo{year}{2023}), \bibinfo{pages}{40853--40871}.
\newblock


\bibitem[Labs(2024)]%
        {flux}
\bibfield{author}{\bibinfo{person}{Black~Forest Labs}.} \bibinfo{year}{2024}\natexlab{}.
\newblock \bibinfo{title}{FLUX}.
\newblock
\newblock
\urldef\tempurl%
\url{https://github.com/black-forest-labs/flux}
\showURL{%
\tempurl}


\bibitem[Lepage et~al\mbox{.}(2023)]%
        {lepage2023lrvsf}
\bibfield{author}{\bibinfo{person}{Simon Lepage}, \bibinfo{person}{Jérémie Mary}, {and} \bibinfo{person}{David Picard}.} \bibinfo{year}{2023}\natexlab{}.
\newblock \showarticletitle{LRVS-Fashion: Extending Visual Search with Referring Instructions}.
\newblock \bibinfo{journal}{\emph{arXiv:2306.02928}} (\bibinfo{year}{2023}).
\newblock


\bibitem[Li et~al\mbox{.}(2024a)]%
        {li2024uv}
\bibfield{author}{\bibinfo{person}{Hong Li}, \bibinfo{person}{Yutang Feng}, \bibinfo{person}{Song Xue}, \bibinfo{person}{Xuhui Liu}, \bibinfo{person}{Bohan Zeng}, \bibinfo{person}{Shanglin Li}, \bibinfo{person}{Boyu Liu}, \bibinfo{person}{Jianzhuang Liu}, \bibinfo{person}{Shumin Han}, {and} \bibinfo{person}{Baochang Zhang}.} \bibinfo{year}{2024}\natexlab{a}.
\newblock \showarticletitle{UV-IDM: Identity-Conditioned Latent Diffusion Model for Face UV-Texture Generation}. In \bibinfo{booktitle}{\emph{Proceedings of the IEEE/CVF Conference on Computer Vision and Pattern Recognition}}. \bibinfo{pages}{10585--10595}.
\newblock


\bibitem[Li et~al\mbox{.}(2024b)]%
        {li2024zone}
\bibfield{author}{\bibinfo{person}{Shanglin Li}, \bibinfo{person}{Bohan Zeng}, \bibinfo{person}{Yutang Feng}, \bibinfo{person}{Sicheng Gao}, \bibinfo{person}{Xiuhui Liu}, \bibinfo{person}{Jiaming Liu}, \bibinfo{person}{Lin Li}, \bibinfo{person}{Xu Tang}, \bibinfo{person}{Yao Hu}, \bibinfo{person}{Jianzhuang Liu}, {et~al\mbox{.}}} \bibinfo{year}{2024}\natexlab{b}.
\newblock \showarticletitle{Zone: Zero-shot instruction-guided local editing}. In \bibinfo{booktitle}{\emph{Proceedings of the IEEE/CVF Conference on Computer Vision and Pattern Recognition}}. \bibinfo{pages}{6254--6263}.
\newblock


\bibitem[Lipman et~al\mbox{.}(2022)]%
        {lipman2022flow}
\bibfield{author}{\bibinfo{person}{Yaron Lipman}, \bibinfo{person}{Ricky~TQ Chen}, \bibinfo{person}{Heli Ben-Hamu}, \bibinfo{person}{Maximilian Nickel}, {and} \bibinfo{person}{Matt Le}.} \bibinfo{year}{2022}\natexlab{}.
\newblock \showarticletitle{Flow matching for generative modeling}.
\newblock \bibinfo{journal}{\emph{arXiv preprint arXiv:2210.02747}} (\bibinfo{year}{2022}).
\newblock


\bibitem[Lu et~al\mbox{.}(2022)]%
        {lu2022dpm}
\bibfield{author}{\bibinfo{person}{Cheng Lu}, \bibinfo{person}{Yuhao Zhou}, \bibinfo{person}{Fan Bao}, \bibinfo{person}{Jianfei Chen}, \bibinfo{person}{Chongxuan Li}, {and} \bibinfo{person}{Jun Zhu}.} \bibinfo{year}{2022}\natexlab{}.
\newblock \showarticletitle{Dpm-solver: A fast ode solver for diffusion probabilistic model sampling in around 10 steps}.
\newblock \bibinfo{journal}{\emph{Advances in Neural Information Processing Systems}}  \bibinfo{volume}{35} (\bibinfo{year}{2022}), \bibinfo{pages}{5775--5787}.
\newblock


\bibitem[Luo et~al\mbox{.}(2023)]%
        {luo2023latent}
\bibfield{author}{\bibinfo{person}{Simian Luo}, \bibinfo{person}{Yiqin Tan}, \bibinfo{person}{Longbo Huang}, \bibinfo{person}{Jian Li}, {and} \bibinfo{person}{Hang Zhao}.} \bibinfo{year}{2023}\natexlab{}.
\newblock \showarticletitle{Latent consistency models: Synthesizing high-resolution images with few-step inference}.
\newblock \bibinfo{journal}{\emph{arXiv preprint arXiv:2310.04378}} (\bibinfo{year}{2023}).
\newblock


\bibitem[Mishchenko and Defazio(2023)]%
        {mishchenko2023prodigy}
\bibfield{author}{\bibinfo{person}{Konstantin Mishchenko} {and} \bibinfo{person}{Aaron Defazio}.} \bibinfo{year}{2023}\natexlab{}.
\newblock \showarticletitle{Prodigy: An expeditiously adaptive parameter-free learner}.
\newblock \bibinfo{journal}{\emph{arXiv preprint arXiv:2306.06101}} (\bibinfo{year}{2023}).
\newblock


\bibitem[Morelli et~al\mbox{.}(2022)]%
        {morelli2022dress}
\bibfield{author}{\bibinfo{person}{Davide Morelli}, \bibinfo{person}{Matteo Fincato}, \bibinfo{person}{Marcella Cornia}, \bibinfo{person}{Federico Landi}, \bibinfo{person}{Fabio Cesari}, {and} \bibinfo{person}{Rita Cucchiara}.} \bibinfo{year}{2022}\natexlab{}.
\newblock \showarticletitle{Dress code: High-resolution multi-category virtual try-on}. In \bibinfo{booktitle}{\emph{Proceedings of the IEEE/CVF conference on computer vision and pattern recognition}}. \bibinfo{pages}{2231--2235}.
\newblock


\bibitem[Peebles and Xie(2023)]%
        {peebles2023scalable}
\bibfield{author}{\bibinfo{person}{William Peebles} {and} \bibinfo{person}{Saining Xie}.} \bibinfo{year}{2023}\natexlab{}.
\newblock \showarticletitle{Scalable diffusion models with transformers}. In \bibinfo{booktitle}{\emph{Proceedings of the IEEE/CVF International Conference on Computer Vision}}. \bibinfo{pages}{4195--4205}.
\newblock


\bibitem[Radford et~al\mbox{.}(2021)]%
        {radford2021learning}
\bibfield{author}{\bibinfo{person}{Alec Radford}, \bibinfo{person}{Jong~Wook Kim}, \bibinfo{person}{Chris Hallacy}, \bibinfo{person}{Aditya Ramesh}, \bibinfo{person}{Gabriel Goh}, \bibinfo{person}{Sandhini Agarwal}, \bibinfo{person}{Girish Sastry}, \bibinfo{person}{Amanda Askell}, \bibinfo{person}{Pamela Mishkin}, \bibinfo{person}{Jack Clark}, {et~al\mbox{.}}} \bibinfo{year}{2021}\natexlab{}.
\newblock \showarticletitle{Learning transferable visual models from natural language supervision}. In \bibinfo{booktitle}{\emph{International conference on machine learning}}. PMLR, \bibinfo{pages}{8748--8763}.
\newblock


\bibitem[Raffel et~al\mbox{.}(2020)]%
        {raffel2020exploring}
\bibfield{author}{\bibinfo{person}{Colin Raffel}, \bibinfo{person}{Noam Shazeer}, \bibinfo{person}{Adam Roberts}, \bibinfo{person}{Katherine Lee}, \bibinfo{person}{Sharan Narang}, \bibinfo{person}{Michael Matena}, \bibinfo{person}{Yanqi Zhou}, \bibinfo{person}{Wei Li}, {and} \bibinfo{person}{Peter~J Liu}.} \bibinfo{year}{2020}\natexlab{}.
\newblock \showarticletitle{Exploring the limits of transfer learning with a unified text-to-text transformer}.
\newblock \bibinfo{journal}{\emph{Journal of machine learning research}} \bibinfo{volume}{21}, \bibinfo{number}{140} (\bibinfo{year}{2020}), \bibinfo{pages}{1--67}.
\newblock


\bibitem[Ravi et~al\mbox{.}(2024)]%
        {ravi2024sam}
\bibfield{author}{\bibinfo{person}{Nikhila Ravi}, \bibinfo{person}{Valentin Gabeur}, \bibinfo{person}{Yuan-Ting Hu}, \bibinfo{person}{Ronghang Hu}, \bibinfo{person}{Chaitanya Ryali}, \bibinfo{person}{Tengyu Ma}, \bibinfo{person}{Haitham Khedr}, \bibinfo{person}{Roman R{\"a}dle}, \bibinfo{person}{Chloe Rolland}, \bibinfo{person}{Laura Gustafson}, {et~al\mbox{.}}} \bibinfo{year}{2024}\natexlab{}.
\newblock \showarticletitle{Sam 2: Segment anything in images and videos}.
\newblock \bibinfo{journal}{\emph{arXiv preprint arXiv:2408.00714}} (\bibinfo{year}{2024}).
\newblock


\bibitem[Rombach et~al\mbox{.}(2022)]%
        {rombach2022high}
\bibfield{author}{\bibinfo{person}{Robin Rombach}, \bibinfo{person}{Andreas Blattmann}, \bibinfo{person}{Dominik Lorenz}, \bibinfo{person}{Patrick Esser}, {and} \bibinfo{person}{Bj{\"o}rn Ommer}.} \bibinfo{year}{2022}\natexlab{}.
\newblock \showarticletitle{High-resolution image synthesis with latent diffusion models}. In \bibinfo{booktitle}{\emph{Proceedings of the IEEE/CVF conference on computer vision and pattern recognition}}. \bibinfo{pages}{10684--10695}.
\newblock


\bibitem[Song et~al\mbox{.}(2020)]%
        {song2020denoising}
\bibfield{author}{\bibinfo{person}{Jiaming Song}, \bibinfo{person}{Chenlin Meng}, {and} \bibinfo{person}{Stefano Ermon}.} \bibinfo{year}{2020}\natexlab{}.
\newblock \showarticletitle{Denoising diffusion implicit models}.
\newblock \bibinfo{journal}{\emph{arXiv preprint arXiv:2010.02502}} (\bibinfo{year}{2020}).
\newblock


\bibitem[Song et~al\mbox{.}(2024a)]%
        {processpainter}
\bibfield{author}{\bibinfo{person}{Yiren Song}, \bibinfo{person}{Shijie Huang}, \bibinfo{person}{Chen Yao}, \bibinfo{person}{Xiaojun Ye}, \bibinfo{person}{Hai Ci}, \bibinfo{person}{Jiaming Liu}, \bibinfo{person}{Yuxuan Zhang}, {and} \bibinfo{person}{Mike~Zheng Shou}.} \bibinfo{year}{2024}\natexlab{a}.
\newblock \showarticletitle{ProcessPainter: Learn Painting Process from Sequence Data}.
\newblock \bibinfo{journal}{\emph{arXiv preprint arXiv:2406.06062}} (\bibinfo{year}{2024}).
\newblock


\bibitem[Song et~al\mbox{.}(2024b)]%
        {song2024diffsimtamingdiffusionmodels}
\bibfield{author}{\bibinfo{person}{Yiren Song}, \bibinfo{person}{Xiaokang Liu}, {and} \bibinfo{person}{Mike~Zheng Shou}.} \bibinfo{year}{2024}\natexlab{b}.
\newblock \bibinfo{title}{DiffSim: Taming Diffusion Models for Evaluating Visual Similarity}.
\newblock
\newblock
\showeprint[arxiv]{2412.14580}~[cs.CV]
\urldef\tempurl%
\url{https://arxiv.org/abs/2412.14580}
\showURL{%
\tempurl}


\bibitem[Song et~al\mbox{.}(2024c)]%
        {anti}
\bibfield{author}{\bibinfo{person}{Yiren Song}, \bibinfo{person}{Shengtao Lou}, \bibinfo{person}{Xiaokang Liu}, \bibinfo{person}{Hai Ci}, \bibinfo{person}{Pei Yang}, \bibinfo{person}{Jiaming Liu}, {and} \bibinfo{person}{Mike~Zheng Shou}.} \bibinfo{year}{2024}\natexlab{c}.
\newblock \showarticletitle{Anti-Reference: Universal and Immediate Defense Against Reference-Based Generation}.
\newblock \bibinfo{journal}{\emph{arXiv preprint arXiv:2412.05980}} (\bibinfo{year}{2024}).
\newblock


\bibitem[Su et~al\mbox{.}(2024)]%
        {su2024roformer}
\bibfield{author}{\bibinfo{person}{Jianlin Su}, \bibinfo{person}{Murtadha Ahmed}, \bibinfo{person}{Yu Lu}, \bibinfo{person}{Shengfeng Pan}, \bibinfo{person}{Wen Bo}, {and} \bibinfo{person}{Yunfeng Liu}.} \bibinfo{year}{2024}\natexlab{}.
\newblock \showarticletitle{Roformer: Enhanced transformer with rotary position embedding}.
\newblock \bibinfo{journal}{\emph{Neurocomputing}}  \bibinfo{volume}{568} (\bibinfo{year}{2024}), \bibinfo{pages}{127063}.
\newblock


\bibitem[Tan et~al\mbox{.}(2024)]%
        {tan2024ominicontrol}
\bibfield{author}{\bibinfo{person}{Zhenxiong Tan}, \bibinfo{person}{Songhua Liu}, \bibinfo{person}{Xingyi Yang}, \bibinfo{person}{Qiaochu Xue}, {and} \bibinfo{person}{Xinchao Wang}.} \bibinfo{year}{2024}\natexlab{}.
\newblock \showarticletitle{OminiControl: Minimal and Universal Control for Diffusion Transformer}.
\newblock \bibinfo{journal}{\emph{arXiv preprint arXiv:2411.15098}} (\bibinfo{year}{2024}).
\newblock


\bibitem[Velioglu et~al\mbox{.}(2024)]%
        {velioglu2024tryoffdiff}
\bibfield{author}{\bibinfo{person}{Riza Velioglu}, \bibinfo{person}{Petra Bevandic}, \bibinfo{person}{Robin Chan}, {and} \bibinfo{person}{Barbara Hammer}.} \bibinfo{year}{2024}\natexlab{}.
\newblock \showarticletitle{TryOffDiff: Virtual-Try-Off via High-Fidelity Garment Reconstruction using Diffusion Models}.
\newblock \bibinfo{journal}{\emph{arXiv preprint arXiv:2411.18350}} (\bibinfo{year}{2024}).
\newblock


\bibitem[Wang et~al\mbox{.}(2024a)]%
        {wang2024instantid}
\bibfield{author}{\bibinfo{person}{Qixun Wang}, \bibinfo{person}{Xu Bai}, \bibinfo{person}{Haofan Wang}, \bibinfo{person}{Zekui Qin}, \bibinfo{person}{Anthony Chen}, \bibinfo{person}{Huaxia Li}, \bibinfo{person}{Xu Tang}, {and} \bibinfo{person}{Yao Hu}.} \bibinfo{year}{2024}\natexlab{a}.
\newblock \showarticletitle{Instantid: Zero-shot identity-preserving generation in seconds}.
\newblock \bibinfo{journal}{\emph{arXiv preprint arXiv:2401.07519}} (\bibinfo{year}{2024}).
\newblock


\bibitem[Wang et~al\mbox{.}(2024b)]%
        {wang2024stablegarment}
\bibfield{author}{\bibinfo{person}{Rui Wang}, \bibinfo{person}{Hailong Guo}, \bibinfo{person}{Jiaming Liu}, \bibinfo{person}{Huaxia Li}, \bibinfo{person}{Haibo Zhao}, \bibinfo{person}{Xu Tang}, \bibinfo{person}{Yao Hu}, \bibinfo{person}{Hao Tang}, {and} \bibinfo{person}{Peipei Li}.} \bibinfo{year}{2024}\natexlab{b}.
\newblock \showarticletitle{StableGarment: Garment-Centric Generation via Stable Diffusion}.
\newblock \bibinfo{journal}{\emph{arXiv preprint arXiv:2403.10783}} (\bibinfo{year}{2024}).
\newblock


\bibitem[Wang et~al\mbox{.}(2004)]%
        {wang2004ssim}
\bibfield{author}{\bibinfo{person}{Zhou Wang}, \bibinfo{person}{Alan~C Bovik}, \bibinfo{person}{Hamid~R Sheikh}, {and} \bibinfo{person}{Eero~P Simoncelli}.} \bibinfo{year}{2004}\natexlab{}.
\newblock \showarticletitle{Image quality assessment: from error visibility to structural similarity}.
\newblock \bibinfo{journal}{\emph{IEEE transactions on image processing}} \bibinfo{volume}{13}, \bibinfo{number}{4} (\bibinfo{year}{2004}), \bibinfo{pages}{600--612}.
\newblock


\bibitem[Xiao et~al\mbox{.}(2024)]%
        {xiao2024florence}
\bibfield{author}{\bibinfo{person}{Bin Xiao}, \bibinfo{person}{Haiping Wu}, \bibinfo{person}{Weijian Xu}, \bibinfo{person}{Xiyang Dai}, \bibinfo{person}{Houdong Hu}, \bibinfo{person}{Yumao Lu}, \bibinfo{person}{Michael Zeng}, \bibinfo{person}{Ce Liu}, {and} \bibinfo{person}{Lu Yuan}.} \bibinfo{year}{2024}\natexlab{}.
\newblock \showarticletitle{Florence-2: Advancing a unified representation for a variety of vision tasks}. In \bibinfo{booktitle}{\emph{Proceedings of the IEEE/CVF Conference on Computer Vision and Pattern Recognition}}. \bibinfo{pages}{4818--4829}.
\newblock


\bibitem[Xie et~al\mbox{.}(2023)]%
        {xie2023gp}
\bibfield{author}{\bibinfo{person}{Zhenyu Xie}, \bibinfo{person}{Zaiyu Huang}, \bibinfo{person}{Xin Dong}, \bibinfo{person}{Fuwei Zhao}, \bibinfo{person}{Haoye Dong}, \bibinfo{person}{Xijin Zhang}, \bibinfo{person}{Feida Zhu}, {and} \bibinfo{person}{Xiaodan Liang}.} \bibinfo{year}{2023}\natexlab{}.
\newblock \showarticletitle{Gp-vton: Towards general purpose virtual try-on via collaborative local-flow global-parsing learning}. In \bibinfo{booktitle}{\emph{Proceedings of the IEEE/CVF Conference on Computer Vision and Pattern Recognition}}. \bibinfo{pages}{23550--23559}.
\newblock


\bibitem[Xu et~al\mbox{.}(2024)]%
        {xu2024ootdiffusion}
\bibfield{author}{\bibinfo{person}{Yuhao Xu}, \bibinfo{person}{Tao Gu}, \bibinfo{person}{Weifeng Chen}, {and} \bibinfo{person}{Chengcai Chen}.} \bibinfo{year}{2024}\natexlab{}.
\newblock \showarticletitle{Ootdiffusion: Outfitting fusion based latent diffusion for controllable virtual try-on}.
\newblock \bibinfo{journal}{\emph{arXiv preprint arXiv:2403.01779}} (\bibinfo{year}{2024}).
\newblock


\bibitem[Ye et~al\mbox{.}(2023)]%
        {ye2023ip}
\bibfield{author}{\bibinfo{person}{Hu Ye}, \bibinfo{person}{Jun Zhang}, \bibinfo{person}{Sibo Liu}, \bibinfo{person}{Xiao Han}, {and} \bibinfo{person}{Wei Yang}.} \bibinfo{year}{2023}\natexlab{}.
\newblock \showarticletitle{Ip-adapter: Text compatible image prompt adapter for text-to-image diffusion models}.
\newblock \bibinfo{journal}{\emph{arXiv preprint arXiv:2308.06721}} (\bibinfo{year}{2023}).
\newblock


\bibitem[Zeng et~al\mbox{.}(2023a)]%
        {zeng2023ipdreamer}
\bibfield{author}{\bibinfo{person}{Bohan Zeng}, \bibinfo{person}{Shanglin Li}, \bibinfo{person}{Yutang Feng}, \bibinfo{person}{Hong Li}, \bibinfo{person}{Sicheng Gao}, \bibinfo{person}{Jiaming Liu}, \bibinfo{person}{Huaxia Li}, \bibinfo{person}{Xu Tang}, \bibinfo{person}{Jianzhuang Liu}, {and} \bibinfo{person}{Baochang Zhang}.} \bibinfo{year}{2023}\natexlab{a}.
\newblock \showarticletitle{Ipdreamer: Appearance-controllable 3d object generation with image prompts}.
\newblock \bibinfo{journal}{\emph{arXiv preprint arXiv:2310.05375}} (\bibinfo{year}{2023}).
\newblock


\bibitem[Zeng et~al\mbox{.}(2024)]%
        {zeng2024controllable}
\bibfield{author}{\bibinfo{person}{Bohan Zeng}, \bibinfo{person}{Shanglin Li}, \bibinfo{person}{Xuhui Liu}, \bibinfo{person}{Sicheng Gao}, \bibinfo{person}{Xiaolong Jiang}, \bibinfo{person}{Xu Tang}, \bibinfo{person}{Yao Hu}, \bibinfo{person}{Jianzhuang Liu}, {and} \bibinfo{person}{Baochang Zhang}.} \bibinfo{year}{2024}\natexlab{}.
\newblock \showarticletitle{Controllable mind visual diffusion model}. In \bibinfo{booktitle}{\emph{Proceedings of the AAAI Conference on Artificial Intelligence}}, Vol.~\bibinfo{volume}{38}. \bibinfo{pages}{6935--6943}.
\newblock


\bibitem[Zeng et~al\mbox{.}(2023b)]%
        {zeng2023face}
\bibfield{author}{\bibinfo{person}{Bohan Zeng}, \bibinfo{person}{Xuhui Liu}, \bibinfo{person}{Sicheng Gao}, \bibinfo{person}{Boyu Liu}, \bibinfo{person}{Hong Li}, \bibinfo{person}{Jianzhuang Liu}, {and} \bibinfo{person}{Baochang Zhang}.} \bibinfo{year}{2023}\natexlab{b}.
\newblock \showarticletitle{Face animation with an attribute-guided diffusion model}. In \bibinfo{booktitle}{\emph{Proceedings of the IEEE/CVF Conference on Computer Vision and Pattern Recognition}}. \bibinfo{pages}{628--637}.
\newblock


\bibitem[Zhang et~al\mbox{.}(2023)]%
        {zhang2023adding}
\bibfield{author}{\bibinfo{person}{Lvmin Zhang}, \bibinfo{person}{Anyi Rao}, {and} \bibinfo{person}{Maneesh Agrawala}.} \bibinfo{year}{2023}\natexlab{}.
\newblock \showarticletitle{Adding conditional control to text-to-image diffusion models}. In \bibinfo{booktitle}{\emph{Proceedings of the IEEE/CVF International Conference on Computer Vision}}. \bibinfo{pages}{3836--3847}.
\newblock


\bibitem[Zhang et~al\mbox{.}(2018)]%
        {zhang2018lpips}
\bibfield{author}{\bibinfo{person}{Richard Zhang}, \bibinfo{person}{Phillip Isola}, \bibinfo{person}{Alexei~A Efros}, \bibinfo{person}{Eli Shechtman}, {and} \bibinfo{person}{Oliver Wang}.} \bibinfo{year}{2018}\natexlab{}.
\newblock \showarticletitle{The unreasonable effectiveness of deep features as a perceptual metric}. In \bibinfo{booktitle}{\emph{Proceedings of the IEEE conference on computer vision and pattern recognition}}. \bibinfo{pages}{586--595}.
\newblock


\bibitem[Zhang et~al\mbox{.}(2024a)]%
        {zhang2024mmtryon}
\bibfield{author}{\bibinfo{person}{Xujie Zhang}, \bibinfo{person}{Ente Lin}, \bibinfo{person}{Xiu Li}, \bibinfo{person}{Yuxuan Luo}, \bibinfo{person}{Michael Kampffmeyer}, \bibinfo{person}{Xin Dong}, {and} \bibinfo{person}{Xiaodan Liang}.} \bibinfo{year}{2024}\natexlab{a}.
\newblock \showarticletitle{MMTryon: Multi-Modal Multi-Reference Control for High-Quality Fashion Generation}.
\newblock \bibinfo{journal}{\emph{arXiv preprint arXiv:2405.00448}} (\bibinfo{year}{2024}).
\newblock


\bibitem[Zhang et~al\mbox{.}(2024b)]%
        {zhang2024ssr}
\bibfield{author}{\bibinfo{person}{Yuxuan Zhang}, \bibinfo{person}{Yiren Song}, \bibinfo{person}{Jiaming Liu}, \bibinfo{person}{Rui Wang}, \bibinfo{person}{Jinpeng Yu}, \bibinfo{person}{Hao Tang}, \bibinfo{person}{Huaxia Li}, \bibinfo{person}{Xu Tang}, \bibinfo{person}{Yao Hu}, \bibinfo{person}{Han Pan}, {et~al\mbox{.}}} \bibinfo{year}{2024}\natexlab{b}.
\newblock \showarticletitle{Ssr-encoder: Encoding selective subject representation for subject-driven generation}. In \bibinfo{booktitle}{\emph{Proceedings of the IEEE/CVF Conference on Computer Vision and Pattern Recognition}}. \bibinfo{pages}{8069--8078}.
\newblock


\bibitem[Zhang et~al\mbox{.}(2024c)]%
        {fast}
\bibfield{author}{\bibinfo{person}{Yuxuan Zhang}, \bibinfo{person}{Yiren Song}, \bibinfo{person}{Jinpeng Yu}, \bibinfo{person}{Han Pan}, {and} \bibinfo{person}{Zhongliang Jing}.} \bibinfo{year}{2024}\natexlab{c}.
\newblock \showarticletitle{Fast Personalized Text to Image Synthesis with Attention Injection}. In \bibinfo{booktitle}{\emph{ICASSP 2024-2024 IEEE International Conference on Acoustics, Speech and Signal Processing (ICASSP)}}. IEEE, \bibinfo{pages}{6195--6199}.
\newblock


\bibitem[Zhang et~al\mbox{.}(2024d)]%
        {makeup}
\bibfield{author}{\bibinfo{person}{Yuxuan Zhang}, \bibinfo{person}{Lifu Wei}, \bibinfo{person}{Qing Zhang}, \bibinfo{person}{Yiren Song}, \bibinfo{person}{Jiaming Liu}, \bibinfo{person}{Huaxia Li}, \bibinfo{person}{Xu Tang}, \bibinfo{person}{Yao Hu}, {and} \bibinfo{person}{Haibo Zhao}.} \bibinfo{year}{2024}\natexlab{d}.
\newblock \showarticletitle{Stable-Makeup: When Real-World Makeup Transfer Meets Diffusion Model}.
\newblock \bibinfo{journal}{\emph{arXiv preprint arXiv:2403.07764}} (\bibinfo{year}{2024}).
\newblock


\bibitem[Zhang et~al\mbox{.}(2024e)]%
        {hair}
\bibfield{author}{\bibinfo{person}{Yuxuan Zhang}, \bibinfo{person}{Qing Zhang}, \bibinfo{person}{Yiren Song}, {and} \bibinfo{person}{Jiaming Liu}.} \bibinfo{year}{2024}\natexlab{e}.
\newblock \showarticletitle{Stable-hair: Real-world hair transfer via diffusion model}.
\newblock \bibinfo{journal}{\emph{arXiv preprint arXiv:2407.14078}} (\bibinfo{year}{2024}).
\newblock


\end{thebibliography}

\clearpage

\appendix

\begin{figure}[!t]
\centering
\includegraphics[width=0.85\columnwidth]{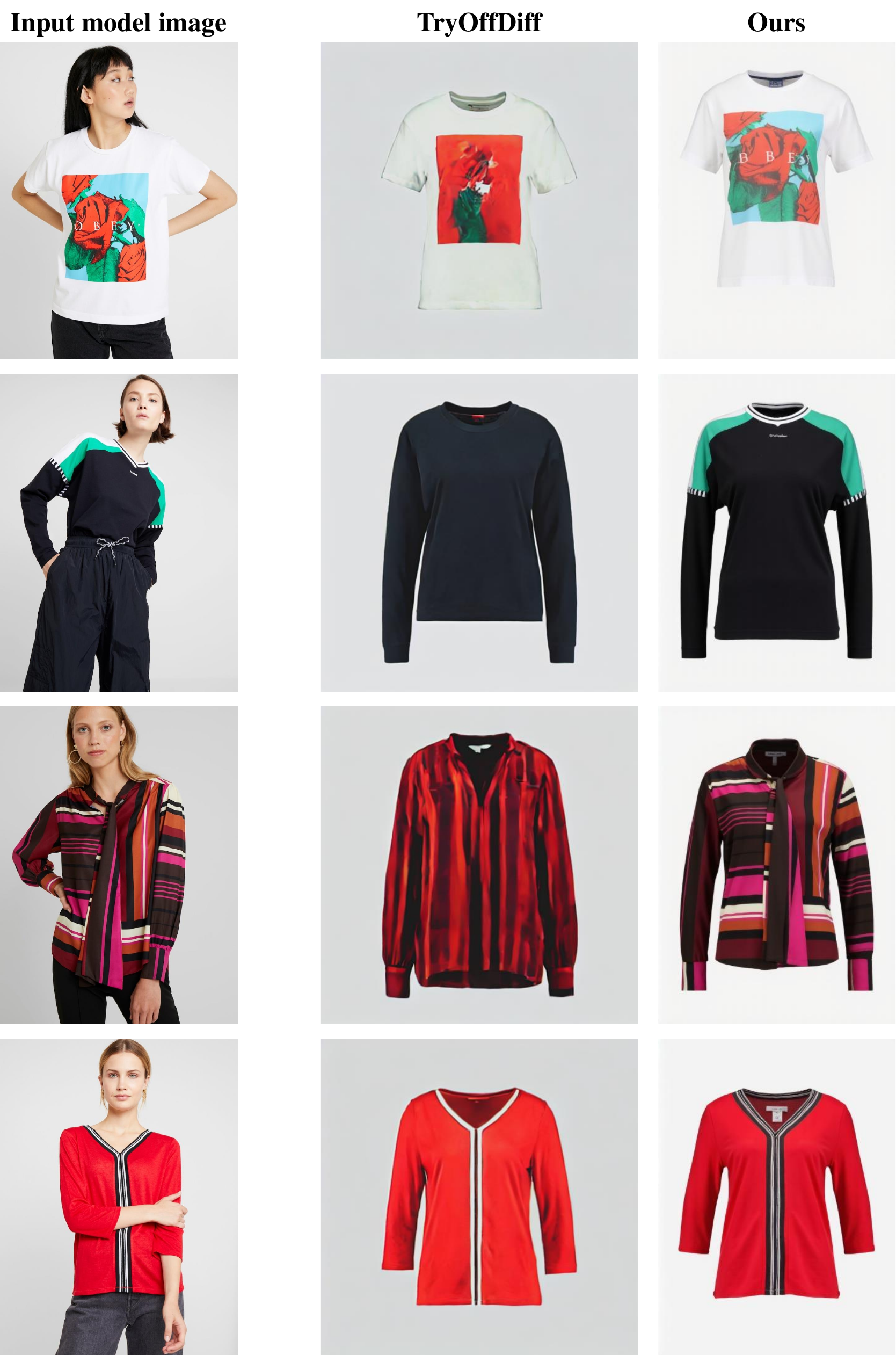}
\caption{Additional qualitative comparison of garment reconstruction.}
\label{Fig:exp_supp_comp_garment}
\end{figure}

\begin{figure}[]
\centering
\includegraphics[width=0.8\columnwidth]{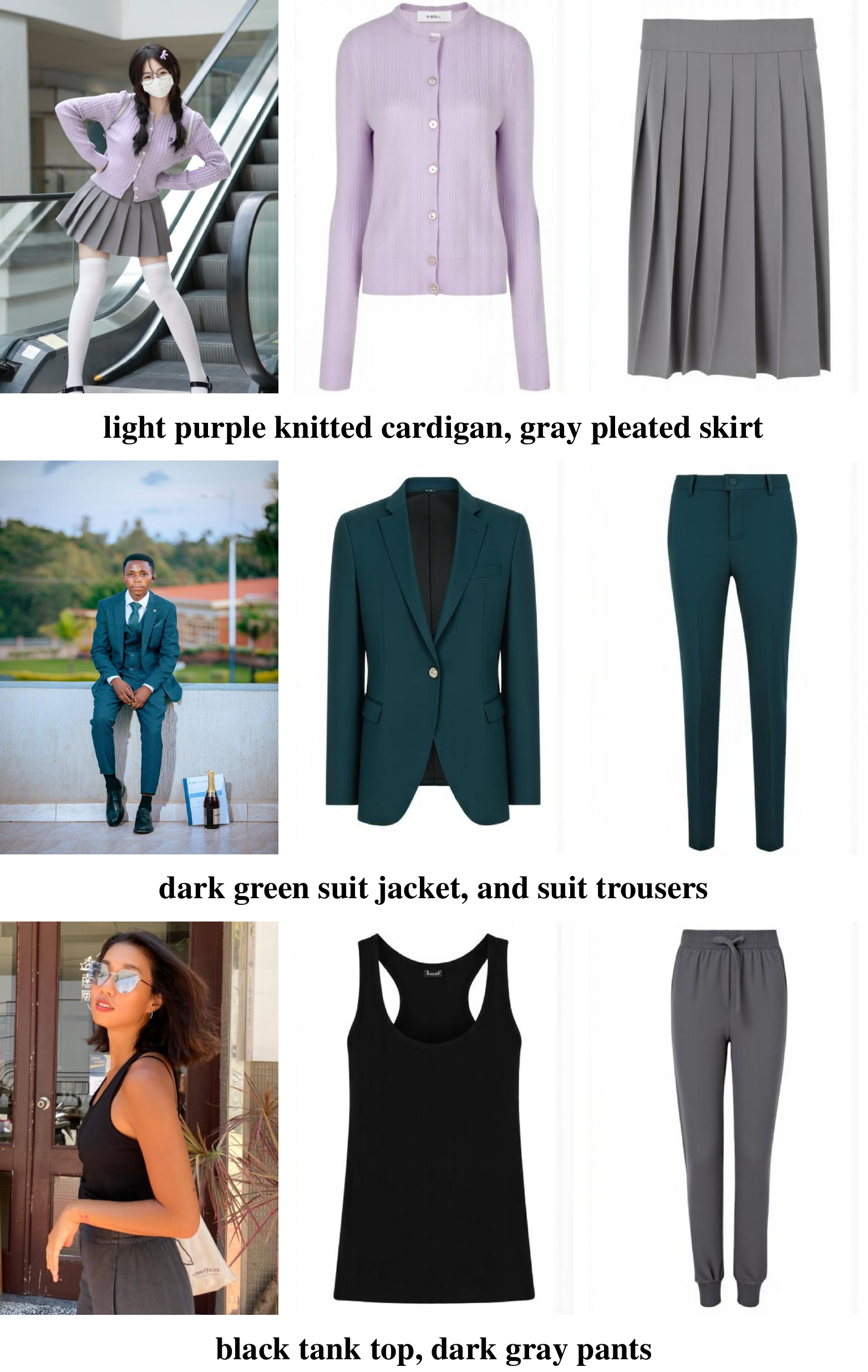}
\caption{Additional garment reconstruction results in the wild.}
\label{Fig:exp_supp_garment_wild}
\end{figure}

\begin{figure*}[]
\centering
\includegraphics[width=0.8\textwidth]{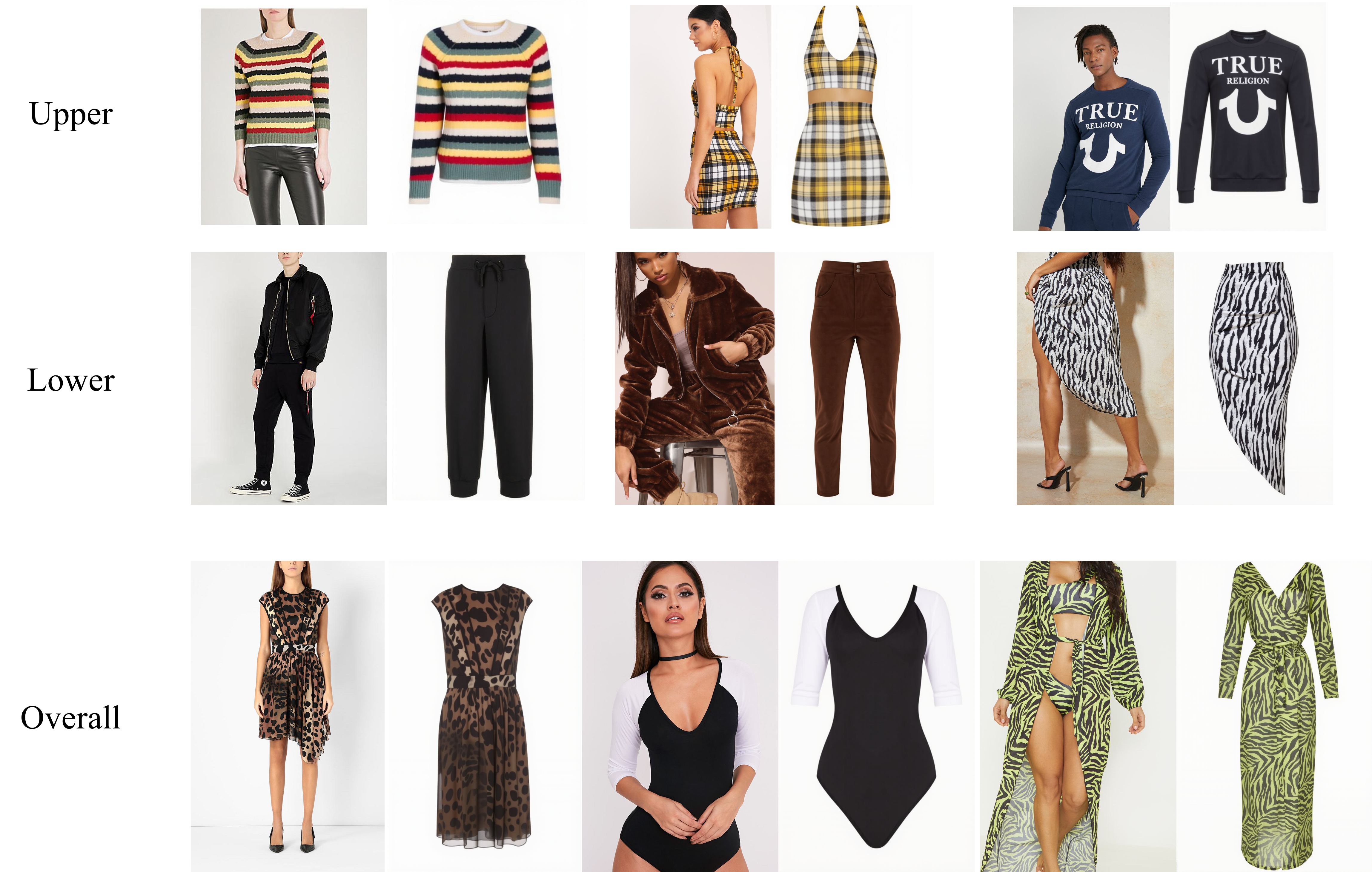}
\caption{Additional garment reconstruction results in the shop.}
\label{Fig:exp_supp_tryoff_shop}
\end{figure*}

\begin{figure*}[]
\centering
\includegraphics[width=0.8\textwidth]{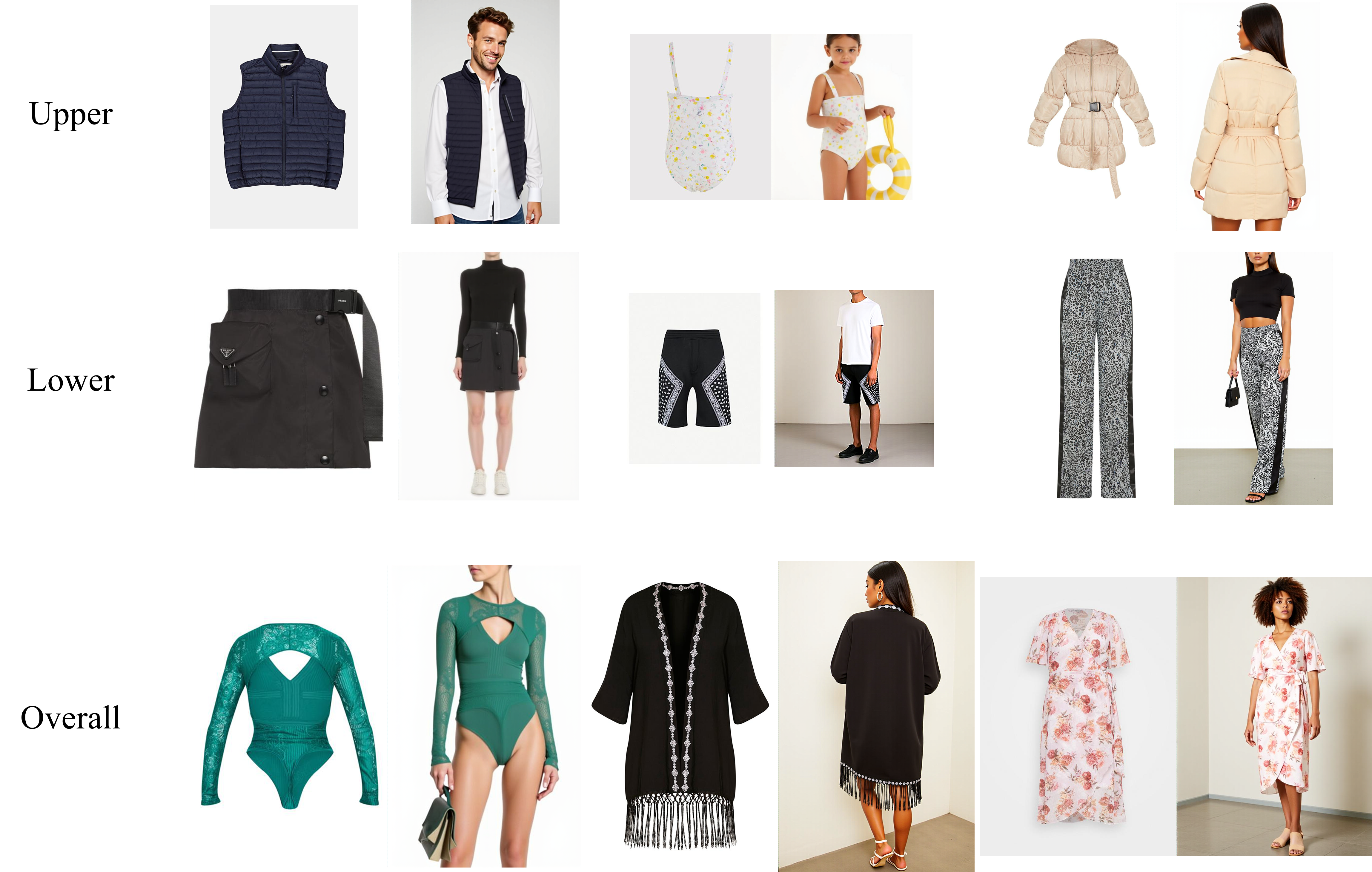}
\caption{Additional model-free virtual try-on results in the shop.}
\label{Fig:exp_supp_modelgen_shop}
\end{figure*}

\begin{figure*}[]
\centering
\includegraphics[width=0.8\textwidth]{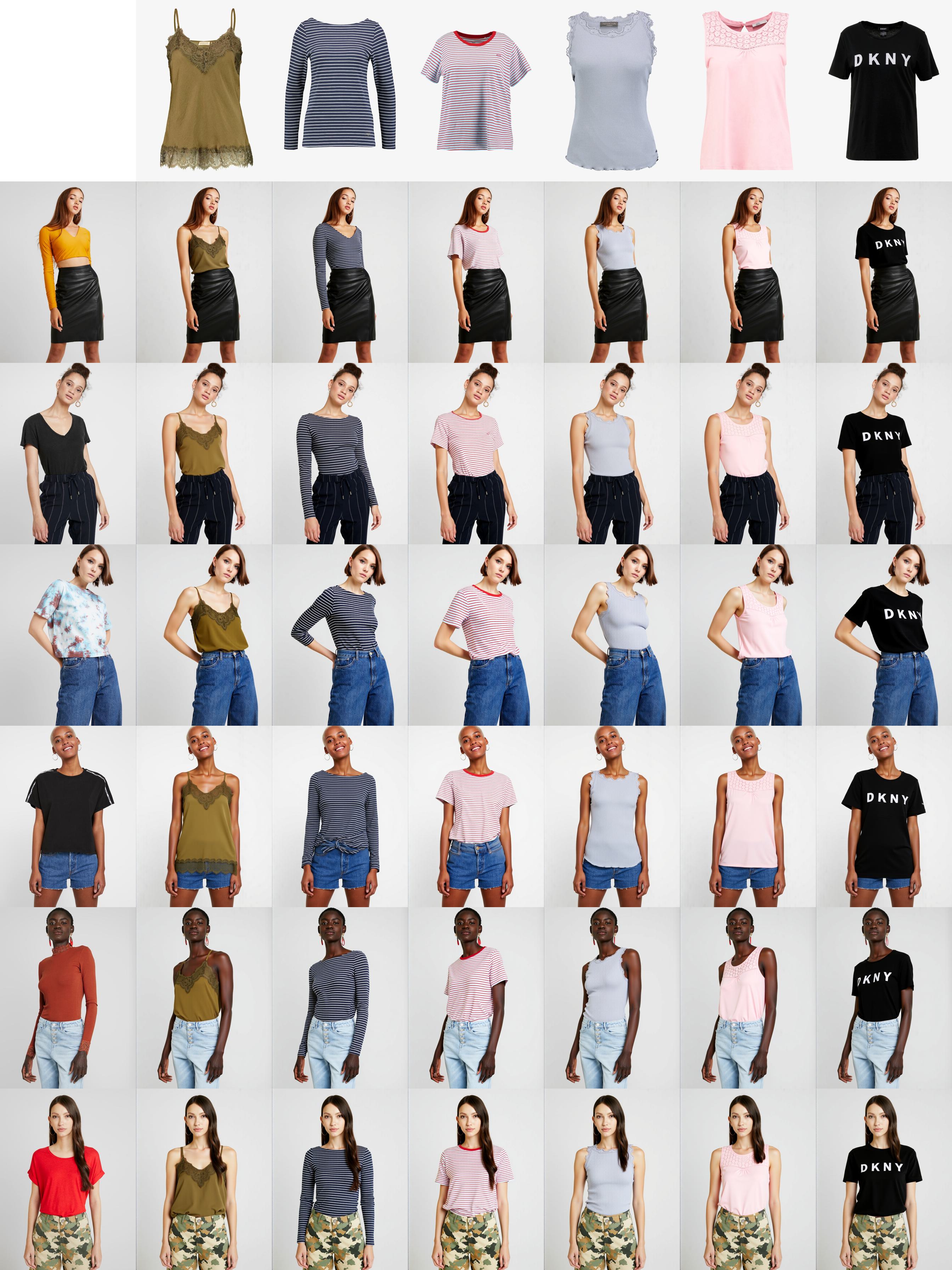}
\caption{Additional virtual try-on results in the shop.}
\label{Fig:exp_supp_tryon_shop}
\end{figure*}

\begin{figure*}[]
\centering
\includegraphics[width=1.8\columnwidth]{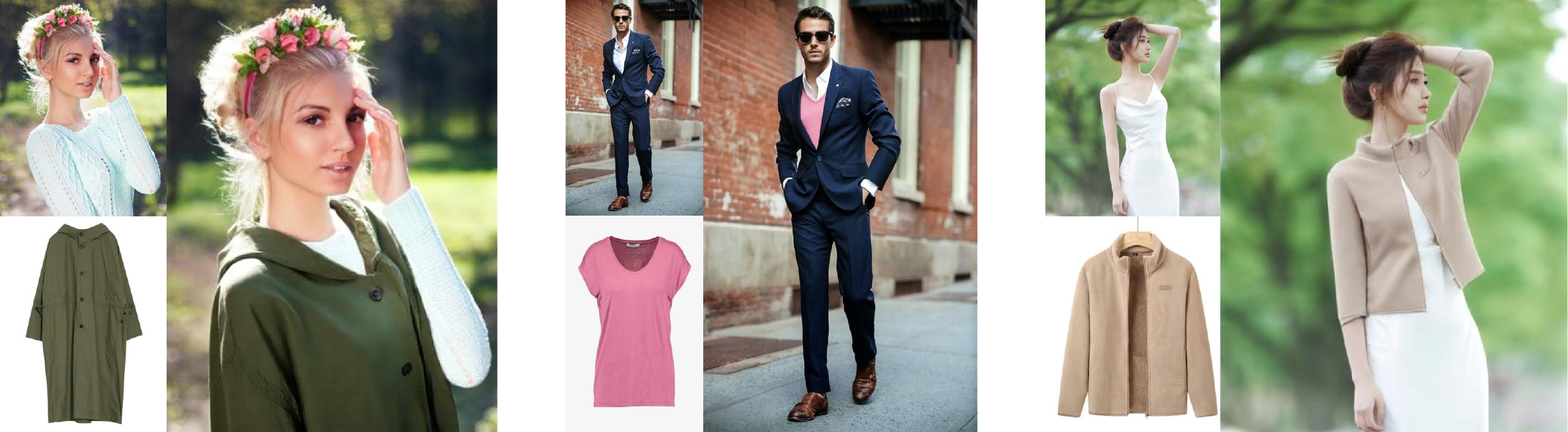}
\caption{Additional VTON in layers results in the wild.}
\label{Fig:exp_supp_try_in_layers_wild}
\end{figure*}

In this Supplementary Material, we provide the details of evaluation metrics in Section.~\ref{supp_sec:metrics}, and in Section.~\ref{supp_sec:more_visual}, we provide more visualization of Any2AnyTryon generation results.

\section{Validation Metrics}
\label{supp_sec:metrics}

\subsection{Garment Reconstruction}
We follow the existing SOTA garment reconstruction method TryOffDiff \cite{velioglu2024tryoffdiff} and leverage the full-reference metrics SSIM~\cite{wang2004ssim}, MS-SSIM, and CW-SSIM to validate the alignment between the reconstructed garment and the ground truth, while utilizing the metrics LPIPS~\cite{zhang2018lpips}, FID~\cite{heusel2017fid}, CLIP-FID, KID~\cite{binkowski2018kid}, and the Deep
 Image Structure and Texture Similarity (DISTS)~\cite{ding2020dists} to evaluate the quality and fidelity of the generated images.

\subsection{Virtual Try-on Genetation}
Our Any2AnyTryon supports both model-free VTON and VTON generation. In the experimental section, we conduct a quantitative comparison of both tasks. For model-free VTON generation, we follow MagiClothing~\cite{chen2024magic} and use MP-LPIPS and CLIP-I to measure the consistency of the garments with the generated outfitted model results. To make our quantitative comparison more compelling, we introduce more recent benchmarks, DiffSim~\cite{song2024diffsimtamingdiffusionmodels} and FFA~\cite{kotar2023these}, to further enhance the validity of the evaluation.

For VTON generation, for paired datasets with ground truth, we use LPIPS~\cite{zhang2018lpips}, SSIM~\cite{wang2004ssim}, FID~\cite{heusel2017fid}, and KID~\cite{binkowski2018kid} to evaluate the quality and faithfulness of the VTON generation. For unpaired datasets without ground truth, we use FID~\cite{heusel2017fid} and KID~\cite{binkowski2018kid} to validate the generation quality.

\section{More VTON Generation Results}
\label{supp_sec:more_visual}

\subsection{Model-free Virtual Try-on}  
In the main paper, we presented a quantitative comparison between Any2AnyTryon and other baseline methods for Model-free Virtual Try-on. To further showcase the high-quality generation results of Any2AnyTryon on the Model-free Virtual Try-on task, we provide additional visualizations of the generated outfitted model images in Fig.~\ref{Fig:exp_supp_modelgen_shop}. To highlight the generated results, we use the user instruction "model in the shop." The results demonstrate that, whether it's for the upper garment, lower garment, or overall outfit change, our method can consistently achieve high-fidelity VTON generation.

\subsection{Virtual Try-on}  
In Fig.~\ref{Fig:exp_supp_tryon_shop}, we display more generation results of Any2AnyTryon in the Virtual Try-on task. We provide six different models and six garments with distinct styles, and our Any2AnyTryon produces 36 different, rational, high-quality generated outfitted results. This proves that our Any2AnyTryon can stably realize impressive VTON generation.

\subsection{Garment Reconstruction}  
To further evaluate the quality of garment reconstruction, we provide additional qualitative comparison results in Fig.~\ref{Fig:exp_supp_comp_garment}. The garments reconstructed using our method in Any2AnyTryon preserve the details of the input models' garments better than those reconstructed by TryOffDiff, besides, we provide more garment generation in Fig.~\ref{Fig:exp_supp_tryoff_shop}. Additionally, to further demonstrate the stability and generalization ability of Any2AnyTryon in garment reconstruction, we display results of garment reconstruction for in-the-wild model images in Fig.~\ref{Fig:exp_supp_garment_wild}. The results show that our method can still generate impressive garment results even for more challenging inputs.

\subsection{Virtual Try-on in Layers}  
For the more challenging task of Virtual Try-on in layers, we present Virtual Try-on in layers generation results for in-the-wild model images in Fig.~\ref{Fig:exp_supp_try_in_layers_wild}. The results demonstrate that, even in complex scenarios such as models in the wild, our Any2AnyTryon can still produce rational, high-quality outfitted model images, proving the effectiveness of our method.

\end{document}